\documentclass{article}

\usepackage[margin=3cm]{geometry}

\usepackage[hidelinks]{hyperref}

\usepackage{amsmath,amssymb,amsfonts}
\usepackage{algorithmic}
\usepackage{graphicx}
\usepackage{textcomp}
\usepackage{soul}
\usepackage{float}
\usepackage{booktabs}
\usepackage{multirow}
\usepackage{subcaption}
\usepackage{array}
\usepackage{wrapfig}
\usepackage[utf8]{inputenc}
\usepackage{color, colortbl}
\usepackage{comment}
\usepackage{balance}
\usepackage[inline]{enumitem}
\usepackage{url}
\usepackage{adjustbox}
\usepackage{amsmath}
\usepackage{fontawesome}

\usepackage{tikz-network}
\usetikzlibrary{decorations.text}
\tikzset{every picture/.style={/utils/exec={\sffamily}}}
\SetLayerDistance{2.75}

\definecolor{other}{HTML}{D9D9D9} \definecolor{starting}{HTML}{FB8072} \definecolor{selected}{HTML}{B3DE69} \definecolor{gold}{HTML}{FFD700} \definecolor{equivalence}{HTML}{BC80BD} 

\definecolor{expanded}{HTML}{80B1D3} \definecolor{expanded1}{HTML}{225EA8} \definecolor{expanded2}{HTML}{1D91C0} \definecolor{expanded3}{HTML}{41B6C4} \definecolor{expanded4}{HTML}{7FCDBB} \definecolor{expanded5}{HTML}{C7E9B4} 

\definecolor{lightyellow}{cmyk}{0,0,0.50,0}
\sethlcolor{lightyellow}
\definecolor{yellow}{cmyk}{0,0,0.50,0}

\newcommand{\STAB}[1]{\begin{tabular}{@{}c@{}}#1\end{tabular}}

\title{Geo-Semantic-Parsing: AI-powered geoparsing\\by traversing semantic knowledge graphs\footnote{Post-print of the article published in Decision Support Systems 136, 2020. Please refer to the published version: \url{doi.org/10.1016/j.dss.2020.113346}}}
\author{Leonardo Nizzoli$^{1,2}$, Marco Avvenuti$^2$, Maurizio Tesconi$^1$, Stefano Cresci$^{1,\dagger}$}
\date{$^1$ Institute of Informatics and Telematics, National Research Council (IIT-CNR)\\ $^2$ Dept. of Information Engineering, University of Pisa\\ $^\dagger$ Corresponding author: \textit{s.cresci@iit.cnr.it}}

\begin{document}
\maketitle

\begin{abstract}
Online social networks convey rich information about geospatial facets of reality. However in most cases, geographic information is not explicit and structured, thus preventing its exploitation in real-time applications. We address this limitation by introducing a novel geoparsing and geotagging technique called Geo-Semantic-Parsing (\texttt{GSP}). \texttt{GSP} identifies location references in free text and extracts the corresponding geographic coordinates. To reach this goal, we employ a semantic annotator to identify relevant portions of the input text and to link them to the corresponding entity in a knowledge graph. Then, we devise and experiment with several efficient strategies for traversing the knowledge graph, thus expanding the available set of information for the geoparsing task. Finally, we exploit all available information for learning a regression model that selects the best entity with which to geotag the input text. We evaluate \texttt{GSP} on a well-known reference dataset including almost 10\textit{k} event-related tweets, achieving $F1=0.66$. We extensively compare our results with those of 2 baselines and 3 state-of-the-art geoparsing techniques, achieving the best performance. On the same dataset, competitors obtain $F1 \leq 0.55$. We conclude by providing in-depth analyses of our results, showing that the overall superior performance of \texttt{GSP} is mainly due to a large improvement in recall, with respect to existing techniques.

\end{abstract}


\section{Introduction}
\label{sec:intro}
Online Social Networks (OSN) are privileged observation channels for understanding the geospatial facets of many real-world phenomena~\cite{kordopatis2017geotagging}. Unfortunately, in most cases OSN content lacks explicit and structured geographic information, as in the case of Twitter, where only a minimal fraction (1\% to 4\%) of messages are natively geotagged~\cite{avvenuti2018gsp}. This shortage of explicit geographic information drastically limits the exploitation of OSN data in geospatial Decision Support Systems (DSS)~\mbox{\cite{keenan2019spatial}}. Conversely, the prompt availability of geotagged content would empower existing systems and would open up the possibility to develop new and better geospatial services and applications~\mbox{\cite{andrienko2020sobigdata, divyaa2019towards}}.
As a practical example of this kind, several social media-based systems have been proposed in recent years for mapping and visualizing situational information in the aftermath of mass disasters -- a task dubbed as \textit{crisis mapping} -- in an effort to augment emergency response~\cite{avvenuti2016impromptu,avvenuti2018crismap}. These systems, however, demand geotagged data to be placed on crisis maps, which in turn imposes to perform the geoparsing task on the majority of social media content. Explicit geographic information is not only needed in early warning~\cite{wu2018disaster, avvenuti2016predictability} and emergency response systems~\cite{avvenuti2014ears,avvenuti2017nowcasting,kumar2019location,de2018taggs,singh2019event}, but also in systems and applications for improving event promotion~\cite{yuan2013and,rehman2020building}, touristic planning~\mbox{\cite{colladon2019using,cresci2014towards,brilhante2015planning}}, healthcare accessibility~\mbox{\cite{li2017designing}}, news aggregation~\cite{mahmud2014home} and verification~\cite{middleton2016geoparsing}. In addition, also other important tasks such as the monitoring of epidemics~\cite{lampos2012nowcasting} and crime prevention~\mbox{\cite{kadar2019public, vomfell2018improving,prietocuriel2020crime}} would benefit from the availability of additional geotagged OSN content, let alone those situations in which geographic information is relevant \textit{per se}, such as in demographic studies~\cite{la2019nationality}.

Given the great importance of geotagged data for 
DSS, much effort has been recently devoted to tasks such as geotagging and geoparsing~\cite{ajao2015survey,zheng2018survey}. In detail, \textit{geotagging} is defined as the generic task of associating geographic coordinates to a given document or to a portion of a document (e.g., a token). Instead, \textit{geoparsing} is a more complex task that can be used to perform geotagging and that involves parsing a text, identifying location mentions and complementing them with their corresponding geographic coordinates~\cite{hu2020harvesting}. There exists also other approaches to geotagging that are not necessarily based on 
free text analysis, such as those based on OSN account information~\cite{zola2019twitter} or on social relationships~\cite{hua2015microblog}.

In this work, we focus on the geoparsing task, and we propose a novel technique called Geo-Semantic-Parsing (\texttt{GSP}). \texttt{GSP} is able to achieve state-of-the-art results
by adopting machine learning and artificial intelligence (AI) techniques to extract geographic information from the rich data contained in semantic knowledge graphs, such as \textit{DBpedia} and \textit{GeoNames}. In particular, in a first step \texttt{GSP} leverages a semantic annotator to identify relevant portions of the input text (i.e., the document to geoparse) and to link them to 
pertinent entities in a reference knowledge graph. Then, it exploits several different strategies to traverse the knowledge graph by navigating links between entities. The result of this second step is an expanded set of candidate entities, that are likely to contain relevant geographic information for the task. Finally, among the geographic information of all retrieved entities, we select those with which to geotag the input document by means of a regression model, that we trained on labeled data. The combination of powerful AI techniques and the rich, structured, interconnected data contained in multiple knowledge graphs allows \texttt{GSP} to achieve $F1 = 0.66$, whereas other state-of-the-art techniques and baselines obtain $F1 \leq 0.55$.

More in detail, one of the reasons why our solution achieves unprecedented results is because it mitigates the problem of \textit{toponymic polysemy} -- that is, the fact that the same toponym can refer to different places according to the context in which it appears\footnote{As an example, at the time of writing \textit{GeoNames} returns 5,331 records for ``Rome'', distributed across all six continents: \url{https://www.geonames.org/search.html?q=rome}.}. The majority of traditional geoparsing techniques adopt heuristics to disambiguate toponyms matched in a gazetteer, a solution that might prove ineffective, especially at world-scale.
As a consequence, the application of such techniques is often constrained to geographically-limited areas, in order to achieve satisfactory performance~\cite{middleton2014b}. Instead, \texttt{GSP} mitigates this issue by performing semantic annotation of the input text -- an operation that intrinsically performs disambiguation of tokens based on their context. In addition, our experimental results demonstrate that the expansion and selection steps of \texttt{GSP} also allow to correct some of the errors made by the semantic annotator.
A second reason for the increased performance of \texttt{GSP} is related to the simplicity of previous approaches to this task. In fact, traditional approaches simply match geographic named entities found in the input text with entries in a gazetteer. Contrarily, our solution is based on powerful techniques (e.g., semantic annotation, regression via gradient boosting decision trees, word and graph embeddings) and information-rich semantic knowledge graphs.
Finally, \texttt{GSP} also has a number of additional advantages over previous techniques: it does not require any explicit geographic information (e.g., GPS coordinates, 
timezones), contrarily to~\cite{dredze2013carmen}; it only exploits text data of input documents (e.g., it does not require any user information or social network topology), contrarily to~\cite{hua2015microblog,de2018taggs}; it processes only one text document at a time (e.g., it does not require all tweets from a user's timeline, or many documents on a given topic), contrarily to~\cite{shen2013linking,li2014effective}; it does not require users to specify a target geographic region, but instead it geoparses places all over the world, contrarily to~\cite{middleton2014b}; by leveraging knowledge graphs, \texttt{GSP} is capable of extracting fine-grained, structured geographic information (e.g., at the level of buildings, cities, counties and regions, countries) similarly to~\cite{dredze2013carmen,halterman2017mordecai}.

\paragraph{Contributions of this work}
Our main contributions can be summarized as follows:
\begin{itemize}
    \item we propose a novel geoparsing technique (\texttt{GSP}),
capable of significantly improving state-of-the-art performance at this task;
    \item to reach our goal, we propose and experiment with several different expansion strategies 
to efficiently traverse a knowledge graph, thus expanding the set of entities to scan for geographic information;
\item we learn a regression model to assign a confidence score to all retrieved entities, in order to select only those providing pertinent geographic information; 
\item we experimentally demonstrate the practical advantage of the design choices on which \texttt{GSP} is rooted. The main improvement brought by \texttt{GSP} is a large boost to the \textit{recall} metric, which we attribute to the optimized expansion strategies previously introduced.
\end{itemize}
Adding to the previous scientific contributions, our solution is also based on state-of-the-art technologies and implementations, for all the necessary steps. In particular, we leverage TagMe~\cite{ferragina2011fast} for semantic annotation and entity linking, and Microsoft's gradient boosting framework LightGBM~\cite{ke2017lightgbm} for learning our regression model, which are currently considered the state-of-the-art for the respective tasks. We also design and experiment with a large number of regressors, some obtained via a process of feature engineering resulting from textual analyses of our documents with FLAIR~\cite{akbik2019flair}, while others directly learned from our data via the use of BERT contextual word embeddings~\cite{devlin2019bert} and \textit{rdf2vec} graph node embeddings~\cite{ristoski2019rdf2vec}.

\paragraph{Roadmap}
The remainder of this paper is organized as 
follows. In Section~\ref{sec:related}, we survey existing works for geoparsing and geotagging of social media content. In Section~\ref{sec:method}, we provide background information, and we introduce the \texttt{GSP} technique. In Sections~\ref{sec:expansion} and~\ref{sec:selection}, we respectively delve into the details of the \textit{expansion} and \textit{selection} steps of \texttt{GSP}, also providing experimental results to support our choices. Then, in Section~\ref{sec:results} we describe our dataset, and we report experimental results of \texttt{GSP} and other techniques for the geoparsing task. We conclude with Section~\ref{sec:discussion} discussing our results, and with Section~\ref{sec:conclusions} summarizing our work and highlighting directions for future research and experimentation.

 \section{Related Works}
\label{sec:related}
A recent survey on location prediction on Twitter~\cite{zheng2018survey} proposed a taxonomy of geotagging and geoparsing techniques according to different tasks. These can be: the prediction of
\begin{enumerate*}[label=(\roman*)]
    \item the locations mentioned in tweets,
    \item the tweet origin location, or
    \item the user home location.
\end{enumerate*}
The remainder of this section adopts the same structure, with a particular focus on works dealing with the prediction of mentioned locations, since our contribution also falls in this category.

\subsection{Mentioned location prediction}
\label{subsec:relmentloc}
The goal of this task is to identify locations mentioned in a text and 
link them to the corresponding geographic coordinates. It has been investigated for a long time on well formatted documents -- like news articles, and researchers identified entity mention variability and ambiguity as the two main challenges of this task. However, mentioned location prediction is even more challenging in OSNs, due to the noisy and short user-generated posts.

The most similar work to our present contribution is our previous attempt at this task, where we proposed a preliminary version of the \texttt{GSP} technique~\cite{avvenuti2018gsp}. 
Similarly to this work, the core idea of~\mbox{\cite{avvenuti2018gsp}} is to
first exploit a semantic annotator to identify relevant portions of the input text, and then 
to parse the corresponding semantic resources in search for possible geographic information. In our previous work, however, we only leveraged a very limited number of semantic resources, actually disregarding many nodes of the available knowledge graphs, that could bring useful information for the geoparsing task, as outlined in Figure~\ref{fig:gsp-expansion}. Moreover, the selection of the geographic information 
to geoparse the input document was carried out by means of a binary classifier, based on a Support Vector Machine (SVM). The thorough, 
AI-driven exploration of the semantic knowledge graphs, together with the accurate selection of informative 
graph nodes
via gradient-boosted regression, thus represent the main differences between our previous and current works. In turn, the profitable exploitation of this additional information significantly improves geoparsing performance, as demonstrated by our experimental results, 
reporting $F1 = 0.66$ \textit{vs} $F1 = 0.55$ of our previous work.

Apart from our previous contribution, the task of mentioned location prediction was traditionally tackled in two steps:  
\begin{enumerate*}[label=(\roman*)]
    \item mentioned location recognition, and
    \item their subsequent disambiguation.
\end{enumerate*}
Mentioned location recognition is generally considered as a special type of Named Entity Recognition (NER) task, and treated accordingly~\cite{zhang2014geocoding}. A few recent works also proposed other specific techniques for identifying location names in texts, outperforming traditional approaches based on NER. As an example, LNEx~\cite{al2018location} learns a statistical language model by applying a skip-gram model to token sequences extracted from gazetteers, which contributes to the accurate and rapid detection of location mentions in texts. The work discussed in~\cite{kumar2019location} applies a classification approach based on word embeddings and on a convolutional neural network for detecting location mentions.
The focus of~\cite{skoumas2016location} is instead posed on the exploitation of spatial relations for location estimation. Authors first employ information extraction methods to identify toponyms and spatial relations in a text. Then, they use expectation maximization to learn models based on spatial probability density functions, and they use these models to infer the location of unknown objects.

Location disambiguation is usually performed by matching the detected location NEs with the entries of a geographic gazetteer (e.g., \emph{GeoNames}\footnote{\url{https://www.geonames.org/}}, \emph{OpenStreetMap}\footnote{\url{https://www.openstreetmap.org/}}). Gazetteers also contain the association between toponyms and their geographic coordinates, thus making it trivial to return the geographic coordinates.
An example of this type of approach is \emph{geoparsepy}\footnote{\url{https://pypi.org/project/geoparsepy/}}, which combines NER techniques for the detection of text chunks, containing location mentions, with heuristics to disambiguate and match the detected NEs with \emph{OpenStreetMap} entries~\cite{middleton2014b,middleton2018location}. 
A similar approach was proposed by Halterman with the \emph{mordecai}\footnote{\url{https://github.com/openeventdata/mordecai}} system~\cite{halterman2017mordecai}. The main difference between \emph{geoparsepy} and \emph{mordecai} is that the latter leverages a deep learning neural network classifier to select best candidate coordinates from the matching entries of the \emph{GeoNames} gazetteer. 
Given that both \emph{geoparsepy} and \emph{mordecai} currently represent state-of-the-art and widely-used geoparsing systems, in our evaluation section we compare our geoparsing results with those benchmarks.

The aforementioned works, as well as our present contribution, propose general-purpose geoparsing techniques that are suitable for application to any standalone textual document (e.g., news articles, tweets, emails, etc.). Other geoparsing systems are instead specifically developed for social media, and they exploit some of the features 
available on specific social networking platforms. For example, the systems discussed in~\cite{li2014effective, shen2013linking} leverage a whole user's timeline to collectively disambiguate toponyms, while~\cite{hua2015microblog} exploits user friendship networks to improve geoparsing accuracy. Similarly, TAGGS~\cite{de2018taggs} enhances location disambiguation in Twitter by employing tweet and user metadata together with other contextual spatial information extracted from additional tweets.

\subsection{Post origin location prediction}
This task targets the prediction of the location from which a post is shared, given its textual content and possible additional information. The simplest approaches to this task scan the textual content of the post and user's profile information in search for geographic clues, and they combine the retrieved geographic information to infer the post origin location.
For example,~\mbox{\cite{laylavi2016multi}} applies scoring and ranking algorithms to data extracted from textual content, users' profile location and place labels for predicting locations at the finest possible granularity. Instead,~\mbox{\cite{williams2017improving}} significantly extends the possible sources of geographic information, adding user's network, external resources and knowledge-bases, and posts related to similar topics. The obtained results are combined to provide the final prediction, following an unsupervised approach.

However, most of the approaches tackle the more challenging scenario of predicting a post origin location in the absence of explicit geographic information. For example,~\cite{zubiaga2017towards} employs several off-the-shelf machine learning classification algorithms (e.g., SVMs, Random Forests, etc.) fed with eight tweet- and metadata-derived features to classify global tweets at the country level. More sophisticated approaches look for content similarities between geotagged and non-geotagged posts. The system proposed in~\cite{paraskevopoulos2016has} estimates the location from which a post was generated by exploiting the similarities in the content between the post and a set of geotagged tweets, as well as their time-evolution characteristics. Similarly,~\mbox{\cite{paule2019fine}} leverages a ranking model to relate a non-geotagged tweet to the most similar geotagged ones, based on the content. Then, it predicts the non-geotagged tweet location by combining the locations of the geotagged tweets, 
using a weighted majority voting algorithm.
Contrarily to~\cite{zubiaga2017towards}, both~\cite{paraskevopoulos2016has,paule2019fine} are capable of providing accurate, fine-grained (i.e., within a city) location estimates. However, their application is typically restricted to geographically-limited areas, whereas~\cite{zubiaga2017towards} can be conveniently applied to classify world tweets.
Finally, \mbox{\cite{luceri2020measurement}} tackles post origin location prediction under the interesting perspective of user privacy. Authors present a deep learning approach to violate user geolocation privacy on Twitter, by predicting her last post origin. To do so, the model leverages previous geotag leakages related to the user itself, and her network neighbours. Notably, the model proves able to violate the privacy of the 60\% of the analyzed users. However, authors also propose defensive strategies, based on data perturbation techniques, to reduce prediction accuracy.

\subsection{User home location prediction}
This task aims to identify users home locations -- a goal that is typically achieved by leveraging a portion of the user's posting history. As a striking example of this kind, in~\cite{singh2019event} user home locations are predicted based on historical locations of the same users, extracted using a Markov model. Similarly to~\cite{zubiaga2017towards}, also the technique recently proposed in~\cite{zola2019twitter} focuses on country-level predictions. The latter method is based on a comparison of frequent word distributions from user timelines with country-based lists of popular Web searches from Google Trends. Given a user,~\cite{zola2019twitter} computes a ranked list of possible home countries, weighted by means of a confidence score obtained via statistical and machine learning methods.

The previously described approaches are solely based on a user's posting history (i.e., its timeline). Instead, another large body of work also leveraged the correlation
between strong connectivity patterns in the social graph and geographic proximity in the real world~\cite{kotzias2016home}. Authors of~\cite{davis2011inferring} define as ``locatable''
those users 
that present geographic information 
allowing their geolocation. Then, given a non-locatable user, their proposed technique iteratively considers reciprocal (i.e., bidirectional) social relationships
to infer the user geolocation from its locatable network neighbours. This simple and preliminary approach is similar to those described in~\cite{laylavi2016multi,williams2017improving} for predicting post origin locations, in that it only combines readily and explicitly available geographic information. A similar approach is proposed in~\cite{kotzias2016home} for efficiently detecting users belonging to a given city. The technique exploits both textual tweet content and Twitter social graph. Instead, a more powerful technique is proposed in~\cite{rodrigues2016exploring}, leveraging a probabilistic approach that jointly models geographic labels and Twitter texts of users, organized in the form of a graph representing the friendship network. In detail, authors use a Markov random field probability model to represent the network, and they ground the learning step on a Markov Chain Monte Carlo simulation, that approximates the posterior probability distribution of the missing geographic user labels.
Finally,~\cite{han2014text} presents an integrated geolocation prediction framework and uses it to investigate what factors impact the prediction accuracy. Authors evaluate a range of feature selection methods to obtain ``location indicative words'', and they investigate the impact of non-geotagged tweets, language and user-declared metadata on user location prediction.
For additional references and in-depth discussions of other user home location methods, we point interested readers to the survey by Ajao \textit{et al.}~\cite{ajao2015survey}, and references therein. \section{Geoparsing documents with \texttt{GSP}}
\label{sec:method}
In 
this section, we first formally define the geoparsing task and how geoparsing techniques are evaluated. Then, we provide the conceptual overview of the \texttt{GSP} technique and the rationale for our design choices.

\subsection{Problem definition}
\label{sec:problem}
Geoparsing involves analyzing a textual document, identifying mentions of known locations, and associating the corresponding geographic coordinates to each mentioned location. Given this formulation, a geoparsing technique is defined as a model $\mathcal{GP}$, such that $\mathcal{GP}(t_{i}) = \mathbf{p}_i$, where $t_{i}$ is the $i$-th document in a collection, and $\mathbf{p}_i = \{p_{i,1}, p_{i,2}, \dots, p_{i,N}\}$ is a set of predicted geographic coordinates $p_{i,k}$, each corresponding to a toponym detected by $\mathcal{GP}$ in $t_{i}$. Ideally, we would want the set of predicted coordinates $\mathbf{p}_i$ to be equal to the set of ground truth coordinates $\mathbf{g}_i$ for $t_{i}$:
\[
\mathbf{p}_i \equiv \mathbf{g}_i \; \Leftrightarrow \; p_{i,k} = g_{i,j} \; \forall \; k,j = 1,\ldots,N
\]
Hence, when evaluating the performance of a geoparsing technique, each $p_{i,k} = g_{i,j}$ is considered as a \textit{true positive} prediction. In practice, two coordinates are considered to be equal if their geographic distance is lower than a certain threshold $\mathcal{T}$. For geoparsing tasks, the most common choice is $\mathcal{T}=100$ miles ($\simeq161$ km)~\cite{zheng2018survey}. However, in our work we adopt a more severe $\mathcal{T}=50$ km ($\simeq31$ miles), which is more suitable for practical applications. In addition to correct predictions, geoparsing techniques can also predict coordinates that do not correspond to any ground truth coordinate: $p_{i,k} \notin \mathbf{g}_i$, thus yielding a type I error (or \textit{false positive}). Similarly, they can fail to predict a ground truth coordinate: $g_{i,j} \notin \mathbf{p}_i$, thus yielding a type II error (or \textit{false negative}). Finally, similarly to information retrieval and entity linking tasks, \textit{true negatives} are typically not considered for the evaluation of geoparsing techniques~\cite{zheng2018survey}.

Given the above definitions, we can count 
true positives (\textit{TP}), false positives (\textit{FP}), and false negatives (\textit{FN}), summarizing the results of the application of a geoparsing technique to a collection of documents:
\begin{align*}
TP &= \sum_i\bigl|\mathbf{p}_i \cap \mathbf{g}_i\bigr|\\
FP &= \sum_i\bigl|\mathbf{p}_i \smallsetminus \mathbf{g}_i\bigr| = \sum_i\Bigl|\{x \; | \; x \in \mathbf{p}_i \; \text{and} \; x \notin \mathbf{g}_i\}\Bigr|\\
FN &= \sum_i\bigl|\mathbf{g}_i \smallsetminus \mathbf{p}_i\bigr| = \sum_i\Bigl|\{x \; | \; x \in \mathbf{g}_i \; \text{and} \; x \notin \mathbf{p}_i\}\Bigr|
\end{align*}
where $\mathbf{a} \smallsetminus \mathbf{b}$ is the set difference between $\mathbf{a}$ and $\mathbf{b}$, and $|\mathbf{a}|$ is the cardinality of set $\mathbf{a}$. In the remainder of this work, geoparsing results are assessed by means of standard evaluation metrics based on \textit{TP}, \textit{FP} and \textit{FN}, such as \textit{precision}, \textit{recall}, and \textit{F1}-score (\textit{F1}).

\subsection{Overview of \texttt{\textup{GSP}}}
\label{sec:gsp-overview}

\begin{figure*}[t]
    \centering
\begin{subfigure}[t]{0.6\textwidth}
        \centering
        \captionsetup[subfigure]{labelformat=empty}
        \resizebox{1\textwidth}{!}{\begin{tikzpicture}[
	mycircle/.style={
	circle,
	draw=black,
	text opacity=1,
	inner sep=0pt,
	minimum size=6.5pt}]
	
\node[mycircle,ultra thin,fill=starting] (sta) {};
	\node[right=0.05cm of sta] (lab1) {\tiny starting node};
	
	\node[mycircle,ultra thin,fill=expanded2,right=0.35cm of lab1] (exp) {};
	\node[right=0.05cm of exp] (lab2) {\tiny related node};
	
	\node[mycircle,ultra thin,fill=selected,right=0.35cm of lab2] (sel) {};
	\node[right=0.05cm of sel] (lab3) {\tiny selected node};
	
	\node[mycircle,ultra thin,fill=other,right=0.35cm of lab3] (oth) {};
	\node[right=0.05cm of oth] (lab4) {\tiny other node};
\end{tikzpicture} }
    \end{subfigure}\\\par\bigskip
\begin{subfigure}[t]{0.39\textwidth}
        \centering
        \resizebox{1\textwidth}{!}{\begin{tikzpicture}[multilayer=3d]
\node[inner sep=10pt] (tweet) at (0,-2)
	{\includegraphics[scale=.1]{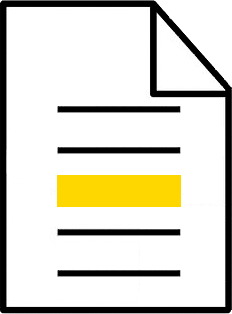}};
	\node (twlabel) at (tweet.south) {$t_i$};
	
\begin{Layer}[layer=1]
		\draw[very thick] (0,0) rectangle (6,3);
		\node at (0,0)[below right]{knowledge graph};
	\end{Layer}
	
\Vertex[x=3,y=1.5,color=starting,layer=1]{A1}
	\Vertex[x=1.5,y=2.5,color=other,label=,layer=1]{B1}
	\Vertex[x=2,y=0.5,color=other,label=,layer=1]{C1}
	\Vertex[x=1,y=0.5,color=other,label=,layer=1]{D1}
	\Vertex[x=4,y=2,color=other,label=,layer=1]{E1}
	\Vertex[x=5,y=1,color=other,label=,layer=1]{F1}
	\Vertex[x=4,y=0.5,color=other,label=,layer=1]{G1}
	\Vertex[x=5,y=2.5,color=other,label=,layer=1]{H1}
	\Vertex[x=1,y=1.5,color=other,label=,layer=1]{I1}

\Edge[](A1)(B1)
	\Edge[](A1)(C1)
	\Edge[](C1)(D1)
	\Edge[](A1)(E1)
	\Edge[](E1)(F1)
	\Edge[](A1)(G1)
	\Edge[](F1)(H1)
	\Edge[](D1)(I1)
	
\draw [->,very thick,color=gold] (tweet.east) to [out=45,in=135] node[] {} (A1.west);
\end{tikzpicture} }
        \caption{Semantic annotation (step 1).}
        \label{fig:gsp-steps-sem}
    \end{subfigure}\begin{subfigure}[t]{0.305\textwidth}
        \centering
        \resizebox{1\textwidth}{!}{\begin{tikzpicture}[multilayer=3d]
\begin{Layer}[layer=1]
		\draw[very thick] (0,0) rectangle (6,3);
		\node at (0,0)[below right]{};
	\end{Layer}
	
\Vertex[x=3,y=1.5,color=starting,layer=1]{A1}
	\Vertex[x=1.5,y=2.5,color=other,label=,layer=1]{B1}
	\Vertex[x=2,y=0.5,color=other,label=,layer=1]{C1}
	\Vertex[x=1,y=0.5,color=other,label=,layer=1]{D1}
	\Vertex[x=4,y=2,color=expanded2,label=,layer=1]{E1}
	\Vertex[x=5,y=1,color=expanded2,label=,layer=1]{F1}
	\Vertex[x=4,y=0.5,color=expanded2,label=,layer=1]{G1}
	\Vertex[x=5,y=2.5,color=expanded2,label=,layer=1]{H1}
	\Vertex[x=1,y=1.5,color=expanded2,label=,layer=1]{I1}

\Edge[](A1)(B1)
	\Edge[](A1)(C1)
	\Edge[](C1)(D1)
	\Edge[](A1)(E1)
	\Edge[](E1)(F1)
	\Edge[](A1)(G1)
	\Edge[](F1)(H1)
	\Edge[](D1)(I1)
\end{tikzpicture} }
        \caption{Expansion (step 2).}
        \label{fig:gsp-steps-exp}
    \end{subfigure}\begin{subfigure}[t]{0.305\textwidth}
        \centering
        \resizebox{1\textwidth}{!}{\begin{tikzpicture}[multilayer=3d]
\begin{Layer}[layer=1]
		\draw[very thick] (0,0) rectangle (6,3);
		\node at (0,0)[below right]{};
	\end{Layer}
	
\Vertex[x=3,y=1.5,color=starting,layer=1]{A1}
	\Vertex[x=1.5,y=2.5,color=other,label=,layer=1]{B1}
	\Vertex[x=2,y=0.5,color=other,label=,layer=1]{C1}
	\Vertex[x=1,y=0.5,color=other,label=,layer=1]{D1}
	\Vertex[x=4,y=2,color=expanded2,label=,layer=1]{E1}
	\Vertex[x=5,y=1,color=expanded2,label=,layer=1]{F1}
	\Vertex[x=4,y=0.5,color=expanded2,label=,layer=1]{G1}
	\Vertex[x=5,y=2.5,color=selected,label=,layer=1]{H1}
	\Vertex[x=1,y=1.5,color=expanded2,label=,layer=1]{I1}

\Edge[](A1)(B1)
	\Edge[](A1)(C1)
	\Edge[](C1)(D1)
	\Edge[](A1)(E1)
	\Edge[](E1)(F1)
	\Edge[](A1)(G1)
	\Edge[](F1)(H1)
	\Edge[](D1)(I1)
\end{tikzpicture} }
        \caption{Selection (step 3).}
        \label{fig:gsp-steps-sel}
    \end{subfigure}\caption{Logical overview of the 3 main steps applied by \texttt{GSP} to the input document $t_i$. Semantic annotation (step 1) links a relevant token (anchor) to an entity (red-colored node) within a reference knowledge graph. Expansion (step 2) identifies related entities (blue-colored nodes) that possibly convey useful geographic information. Selection (step 3) picks the best entity (green-colored node) to geotag the anchor.}
\label{fig:gsp-steps}
\end{figure*}
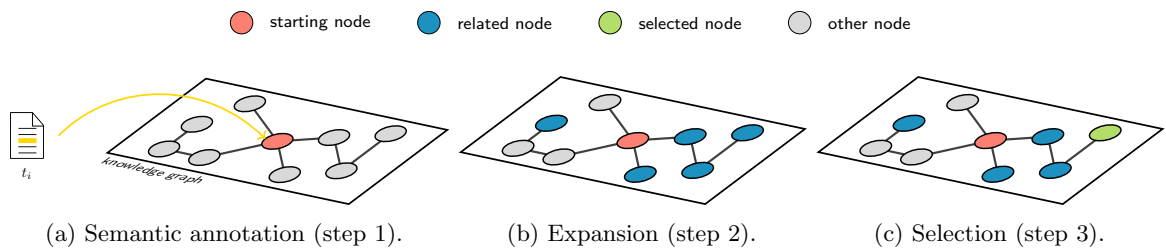

In Figure~\mbox{\ref{fig:gsp-steps}}, we provide a schema of the \texttt{GSP} system. As introduced in Section~\mbox{\ref{sec:intro}}, the main idea behind \texttt{GSP} is to leverage the rich, structured and linked information exposed by a knowledge graph to identify, disambiguate and geotag mentioned locations. To do so, \texttt{GSP} processes a single document \mbox{$t_{i}$} at a time, through three sequential steps:
\begin{itemize}
    \item \textbf{step 1:} \emph{semantic annotation} (Figure~\mbox{\ref{fig:gsp-steps-sem}}), which identifies a relevant token (\emph{anchor}) in the input text \mbox{$t_{i}$} (yellow-colored) and links it to a pertinent entity (red-colored) in a knowledge graph. The purpose of this first step is to augment the input text with the information exposed by the knowledge graph;
    \item \textbf{step 2:} \emph{expansion} (Figure~\mbox{\ref{fig:gsp-steps-exp}}), which traverses the information-rich, structured knowledge graph, retrieving entities (blue-colored) related to the starting one and likely to convey further useful geographic information. The purpose of this intermediate step is to take full advantage of the knowledge graph structure for enriching the available information, thus potentially increasing the model \textit{recall};
    \item \textbf{step 3:} \emph{selection} (Figure~\mbox{\ref{fig:gsp-steps-sel}}), which analyses the entities retrieved by the expansion step to pick the best candidate (green-colored) for geotagging the anchor. In particular, \texttt{GSP} parses the selected entity to extract the geographic coordinates, returned as the final result of the geoparsing process. The purpose of this final step is to deal with the information overload, introduced by the expansion step, thus improving the model \textit{precision}.
\end{itemize}

In the remainder of this Section, we describe and motivate the three steps of \texttt{GSP}. Furthermore, our core scientific contributions -- that is, the \textit{expansion} and \textit{selection} steps -- are also thoroughly discussed and evaluated in Sections~\mbox{\ref{sec:expansion}} and~\mbox{\ref{sec:selection}}. For simplicity, throughout this work we use the terms \textit{knowledge graph} and \textit{knowledge-base} interchangeably, since we always leverage the Linked Data representation of all mentioned knowledge-bases. Similarly, we use the terms \textit{entity} and \textit{resource} interchangeably, when referring to a node of a knowledge graph.

\subsubsection{Step 1: semantic annotation}
\label{sec:method-annotation}
In \texttt{GSP}, we delegate the identification and disambiguation of location mentions to 
semantic annotation, which thus constitutes the starting point of our procedure (step 1 in Figure~\mbox{\ref{fig:gsp-steps}}).
Semantic annotation is a long-studied task for augmenting documents, so that mentions of relevant entities in a text (e.g., persons, places, organizations) are linked to the corresponding entity in a reference knowledge-base~\cite{ferragina2011fast}. This annotation process is highly informative, since it enables the exploitation of the rich information contained in the knowledge-base. By giving access to a wealth of structured and interconnected information, semantic annotation also effectively mitigates the drawbacks related to the sparsity of short social media texts. In addition, it also has the side effect of alleviating possible geoparsing mistakes caused by toponymic polysemy, since semantic annotators automatically carry out disambiguation and only return the most likely reference to a knowledge-base entity for every annotated token. Notably, this disambiguation operation is more accurate than those carried out in previous works, such as those based on simple heuristics as~\cite{middleton2014b}. Semantic annotation is also more powerful than traditional NER for identifying relevant portions of a text, since it gives access to the information of a knowledge-base. Downstream of the annotation step, each relevant anchor of each input text \mbox{$t_{i}$} is linked to the most pertinent entity of a reference knowledge graph, providing the information needed to identify and geotag the mentioned locations. Those entities constitute the input of the subsequent expansion step.

\paragraph{Implementation notes} We perform semantic annotation with TagMe\footnote{\url{https://tagme.d4science.org/tagme/}} -- one of the most popular and best-performing off-the-shelf annotators currently available. In~\cite{avvenuti2018gsp}, we provided geoparsing results comparing the performance of 4 different semantic annotators, with TagMe achieving the best results. Moreover, TagMe is particularly suitable for our work, since it is specifically designed for short and poorly written texts, such as social media messages~\cite{ferragina2011fast}. In order to have complete access to all its functionalities and to allow fast queries, we leverage a local deployment~\cite{hasibi2016reproducibility}. By default, TagMe annotates documents with \textit{Wikipedia} entities. However, for each such entity we refer to its equivalent on \textit{DBpedia}, in order to exploit Linked Data properties and relations. Thus, (the English) \textit{DBpedia} is our reference knowledge graph.

\subsubsection{Step 2: expansion}
\label{sec:method-expansion}

\begin{figure*}[t]
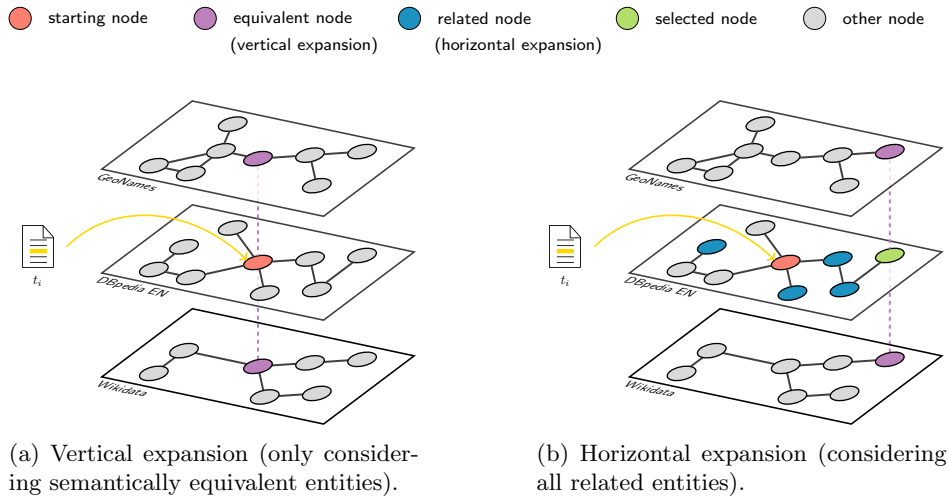

    \centering
\begin{subfigure}[t]{0.8\textwidth}
        \centering
        \captionsetup[subfigure]{labelformat=empty}
        \resizebox{1\textwidth}{!}{\begin{tikzpicture}[
	mycircle/.style={
	circle,
	draw=black,
	text opacity=1,
	inner sep=0pt,
	minimum size=6.5pt}]
	
\node[mycircle,ultra thin,fill=starting] (sta) {};
	\node[right=0.05cm of sta] (lab1) {\tiny starting node};
	
	\node[mycircle,ultra thin,fill=equivalence,right=0.35cm of lab1] (equ) {};
	\node[right=0.05cm of equ] (lab2) {\tiny equivalent node};
	\node[below right=-0.01cm and 0.05cm of equ] (lab2bis) {\tiny (vertical expansion)};
	
	\node[mycircle,ultra thin,fill=expanded2,right=0.35cm of lab2] (exp) {};
	\node[right=0.05cm of exp] (lab3) {\tiny related node};
	\node[below right=-0.01cm and 0.05cm of exp] (lab3bis) {\tiny (horizontal expansion)};
	
	\node[mycircle,ultra thin,fill=selected,right=0.75cm of lab3] (sel) {};
	\node[right=0.05cm of sel] (lab4) {\tiny selected node};
	
	\node[mycircle,ultra thin,fill=other,right=0.35cm of lab4] (oth) {};
	\node[right=0.05cm of oth] (lab5) {\tiny other node};
\end{tikzpicture} }
    \end{subfigure}\\\par\bigskip
\begin{subfigure}[t]{0.35\textwidth}
        \centering
        \resizebox{1\textwidth}{!}{\begin{tikzpicture}[multilayer=3d]
\node[inner sep=10pt] (tweet) at (4,-5)
	{\includegraphics[scale=.1]{images/tweet-document-selected.png}};
	\node (twlabel) at (tweet.south) {$t_i$};
	
\begin{Layer}[layer=1]
		\draw[very thick] (0,0) rectangle (6,3);
		\node at (0,0)[below right]{Wikidata};
	\end{Layer}
\Vertex[x=3,y=1.5,color=equivalence,layer=1]{A1}
	\Vertex[x=1,y=0.5,color=other,label=,layer=1]{D1}
	\Vertex[x=4,y=2,color=other,label=,layer=1]{E1}
	\Vertex[x=5,y=1,color=other,label=,layer=1]{F1}
	\Vertex[x=4,y=0.5,color=other,label=,layer=1]{G1}
	\Vertex[x=5,y=2.5,color=other,label=,layer=1]{H1}
	\Vertex[x=1,y=1.5,color=other,label=,layer=1]{I1}
\Edge[](A1)(I1)
    \Edge[](I1)(D1)
    \Edge[](A1)(G1)
    \Edge[](G1)(F1)
    \Edge[](A1)(E1)
    \Edge[](E1)(H1)
	
\begin{Layer}[layer=2]
		\draw[very thick,fill=white,opacity=0.75] (0,0) rectangle (6,3);
		\node at (0,0)[below right]{DBpedia EN};
	\end{Layer}
\Vertex[x=3,y=1.5,color=starting,layer=2]{A2}
	\Vertex[x=1.5,y=2.5,color=other,label=,layer=2]{B2}
	\Vertex[x=2,y=0.5,color=other,label=,layer=2]{C2}
	\Vertex[x=1,y=0.5,color=other,label=,layer=2]{D2}
	\Vertex[x=4,y=2,color=other,label=,layer=2]{E2}
	\Vertex[x=5,y=1,color=other,label=,layer=2]{F2}
	\Vertex[x=4,y=0.5,color=other,label=,layer=2]{G2}
	\Vertex[x=5,y=2.5,color=other,label=,layer=2]{H2}
	\Vertex[x=1,y=1.5,color=other,label=,layer=2]{I2}
\Edge[NotInBG](A2)(B2)
	\Edge[NotInBG](A2)(C2)
	\Edge[NotInBG](C2)(D2)
	\Edge[NotInBG](A2)(E2)
	\Edge[NotInBG](E2)(F2)
	\Edge[NotInBG](A2)(G2)
	\Edge[NotInBG](F2)(H2)
	\Edge[NotInBG](D2)(I2)
	
\Vertex[x=3,y=1.5,shape=coordinate,layer=3]{A3}
\Edge[style=dashed,color=equivalence](A1)(A2)
	\Edge[style=dashed,color=equivalence,NotInBG](A3)(A2)
	
\begin{Layer}[layer=3]
		\draw[very thick,fill=white,opacity=0.75] (0,0) rectangle (6,3);
		\node at (0,0)[below right]{GeoNames};
	\end{Layer}	
\Vertex[x=3,y=1.5,color=equivalence,layer=3]{A3}
	\Vertex[x=1.5,y=2.5,color=other,label=,layer=3]{B3}
	\Vertex[x=2,y=0.5,color=other,label=,layer=3]{C3}
	\Vertex[x=1,y=0.5,color=other,label=,layer=3]{D3}
	\Vertex[x=4,y=2,color=other,label=,layer=3]{E3}
	\Vertex[x=5,y=1,color=other,label=,layer=3]{F3}
	\Vertex[x=5,y=2.5,color=other,label=,layer=3]{H3}
	\Vertex[x=2,y=1.5,color=other,label=,layer=3]{L3}
\Edge[NotInBG](A3)(L3)
	\Edge[NotInBG](L3)(C3)
	\Edge[NotInBG](C3)(D3)
	\Edge[NotInBG](D3)(L3)
	\Edge[NotInBG](L3)(B3)
	\Edge[NotInBG](A3)(E3)
	\Edge[NotInBG](E3)(H3)
	\Edge[NotInBG](E3)(F3)
	
\draw [->,very thick,color=gold] (tweet.east) to [out=45,in=135] node[] {} (A2.west);
\end{tikzpicture} }
        \caption{Vertical expansion (only considering semantically equivalent entities).}
        \label{fig:gsp-exp-old}
    \end{subfigure}\hspace{0.1\textwidth}\begin{subfigure}[t]{0.35\textwidth}
        \centering
        \resizebox{1\textwidth}{!}{\begin{tikzpicture}[multilayer=3d]
\node[inner sep=10pt] (tweet) at (4,-5)
	{\includegraphics[scale=.1]{images/tweet-document-selected.png}};
	\node (twlabel) at (tweet.south) {$t_i$};
	
\begin{Layer}[layer=1]
		\draw[very thick] (0,0) rectangle (6,3);
		\node at (0,0)[below right]{Wikidata};
	\end{Layer}
\Vertex[x=3,y=1.5,color=other,layer=1]{A1}
	\Vertex[x=1,y=0.5,color=other,label=,layer=1]{D1}
	\Vertex[x=4,y=2,color=other,label=,layer=1]{E1}
	\Vertex[x=5,y=1,color=other,label=,layer=1]{F1}
	\Vertex[x=4,y=0.5,color=other,label=,layer=1]{G1}
	\Vertex[x=5,y=2.5,color=equivalence,label=,layer=1]{H1}
	\Vertex[x=1,y=1.5,color=other,label=,layer=1]{I1}
\Edge[](A1)(I1)
    \Edge[](I1)(D1)
    \Edge[](A1)(G1)
    \Edge[](G1)(F1)
    \Edge[](A1)(E1)
    \Edge[](E1)(H1)
	
\begin{Layer}[layer=2]
		\draw[very thick,fill=white,opacity=0.75] (0,0) rectangle (6,3);
		\node at (0,0)[below right]{DBpedia EN};
	\end{Layer}
\Vertex[x=3,y=1.5,color=starting,layer=2]{A2}
	\Vertex[x=1.5,y=2.5,color=other,label=,layer=2]{B2}
	\Vertex[x=2,y=0.5,color=other,label=,layer=2]{C2}
	\Vertex[x=1,y=0.5,color=other,label=,layer=2]{D2}
	\Vertex[x=4,y=2,color=expanded2,label=,layer=2]{E2}
	\Vertex[x=5,y=1,color=expanded2,label=,layer=2]{F2}
	\Vertex[x=4,y=0.5,color=expanded2,label=,layer=2]{G2}
	\Vertex[x=5,y=2.5,color=selected,label=,layer=2]{H2}
	\Vertex[x=1,y=1.5,color=expanded2,label=,layer=2]{I2}
\Edge[NotInBG](A2)(B2)
	\Edge[NotInBG](A2)(C2)
	\Edge[NotInBG](C2)(D2)
	\Edge[NotInBG](A2)(E2)
	\Edge[NotInBG](E2)(F2)
	\Edge[NotInBG](A2)(G2)
	\Edge[NotInBG](F2)(H2)
	\Edge[NotInBG](D2)(I2)
	
\Vertex[x=5,y=2.5,shape=coordinate,layer=3]{H3}
\Edge[style=dashed,color=equivalence](H1)(H2)
	\Edge[style=dashed,color=equivalence,NotInBG](H3)(H2)
	
\begin{Layer}[layer=3]
		\draw[very thick,fill=white,opacity=0.75] (0,0) rectangle (6,3);
		\node at (0,0)[below right]{GeoNames};
	\end{Layer}	
\Vertex[x=3,y=1.5,color=other,layer=3]{A3}
	\Vertex[x=1.5,y=2.5,color=other,label=,layer=3]{B3}
	\Vertex[x=2,y=0.5,color=other,label=,layer=3]{C3}
	\Vertex[x=1,y=0.5,color=other,label=,layer=3]{D3}
	\Vertex[x=4,y=2,color=other,label=,layer=3]{E3}
	\Vertex[x=5,y=1,color=other,label=,layer=3]{F3}
	\Vertex[x=5,y=2.5,color=equivalence,label=,layer=3]{H3}
	\Vertex[x=2,y=1.5,color=other,label=,layer=3]{L3}
\Edge[NotInBG](A3)(L3)
	\Edge[NotInBG](L3)(C3)
	\Edge[NotInBG](C3)(D3)
	\Edge[NotInBG](D3)(L3)
	\Edge[NotInBG](L3)(B3)
	\Edge[NotInBG](A3)(E3)
	\Edge[NotInBG](E3)(H3)
	\Edge[NotInBG](E3)(F3)
	
\draw [->,very thick,color=gold] (tweet.east) to [out=45,in=135] node[] {} (A2.west);
\end{tikzpicture} }
        \caption{Horizontal expansion (considering all related entities).}
        \label{fig:gsp-exp-new}
    \end{subfigure}
    \caption{Difference between vertical and horizontal expansion. Vertical expansion traverses semantically equivalent entities (purple-colored) across different knowledge graphs, whereas horizontal expansion considers those nodes that are most related (blue-colored) to the starting one, within the reference knowledge graph.}
\label{fig:gsp-expansion}
\end{figure*}

As a result of the semantic annotation step, we have documents where relevant tokens are identified and linked to entities in a knowledge graph. By parsing the structured information associated with these entities, we would thus be able to retrieve the coordinates of geographic entities, whenever available, thus solving the geoparsing task. However, this na\"{i}ve approach would only exploit information of a single node in a graph, thus negating the advantage of knowledge graphs in the first place and ignoring many other potentially informative nodes. Instead, being able to combine information from multiple entities would allow to correct wrong or missing information on a node\footnote{We recall that Linked Data knowledge graphs are collaboratively curated. As with all user-generated content, mistakes and inconsistencies are indeed possible and should be accounted for.}. Furthermore, it would also allow to correct at least part of the errors resulting from the semantic annotation step.

Because of these reasons, both our earlier~\cite{avvenuti2018gsp} and our present geoparsing technique perform an \textit{expansion} step. Given a starting node (i.e., the one linked to a token in $t_i$ by the semantic annotator), the goal of this second step is 
to find other nodes that are related to the starting one, within which to look for relevant information for the geoparsing task, as sketched in Figure~\mbox{\ref{fig:gsp-steps-exp}}. To perform expansion, in~\cite{avvenuti2018gsp} we exploited relations of semantic equivalence (i.e., \texttt{owl:sameAs} links) between entities. These links connect representations of the same entity across different knowledge-bases. Thus, in order to extract geographic information about an entity, in~\cite{avvenuti2018gsp} we also exploited information about all semantically-equivalent entities that are reachable by iteratively navigating equivalence links. As shown in Figure~\ref{fig:gsp-exp-old} via the formalism of multilayer networks, this expansion unfolds in a visually \textit{vertical} fashion. On the one hand, it allows to leverage information coming from more than one node. On the other hand, however, the total number of nodes reachable via this expansion strategy ($\approx 10^1$) is still underwhelming, when compared to the total number of nodes available in a knowledge graph ($\approx 10^6-10^7$). Moreover, it only exploits relations of semantic equivalence, disregarding all the other types of relations between entities.

In our new \texttt{GSP} technique, we greatly increase the number of nodes that we leverage for geoparsing. We reach this goal by devising and experimenting with several different expansion strategies that, given a starting node, are capable of retrieving a large (i.e., with potentially hundreds of entities) ordered vector of related -- but not equivalent -- nodes from within the same knowledge graph. The common idea to all these expansion strategies is to explore nodes \textit{horizontally} rather than vertically (i.e., within a knowledge graph \textit{vs} across knowledge graphs), as shown in Figure~\ref{fig:gsp-exp-new}. This approach has the advantage of retrieving many more nodes (together with all their associated information) with respect to that of~\cite{avvenuti2018gsp}. In turn, this boosts \texttt{GSP}'s \textit{recall} -- that is, the capacity of extracting geographic coordinates for toponyms in the input document. However, retrieving too many, possibly unrelated nodes 
can impair \textit{precision}. Because of this trade-off, it is important to 
evaluate different horizontal expansion strategies, as extensively investigated in Section~\ref{sec:expansion}.

\paragraph{Implementation notes} All our horizontal expansion strategies accept a configurable size parameter $L$, representing the number of nodes to retrieve. In fact, contrarily to the vertical expansion used in~\cite{avvenuti2018gsp}, horizontal expansions are ``unconstrained'' and can potentially return all nodes in a graph. Therefore, in the remainder we refer to \textsf{strategy}$\rvert_{L}$, meaning a specific expansion strategy and the number $L$ of nodes it returns. 
Moreover, all expansion strategies only return nodes with geographic information. Nodes that are not complemented with any geographic information are not considered during expansion, even if related or similar to the starting node, since they could not be used for geoparsing anyway.
Finally, we constrain expansion strategies to depend solely on the starting node returned by the semantic annotator. In this way, the expansions can be pre-computed once and for all for the entire knowledge graph, thus avoiding to perform this demanding operation at runtime.

\subsubsection{Step 3: selection}
\label{sec:method-selection}
After the expansion step, we have access to a potentially large set of candidate nodes. Intuitively, the better is the expansion, the easier it is the selection of the node from which to extract geographic information. In fact, if the expansion only returned entities that are strictly geographically-related to the starting one, then any of those entities would provide pertinent geographic information for the geoparsing task, thus rendering the selection step trivial. However, in the majority of cases, the expansion step also provides some unrelated entities that should not be considered for geoparsing. Thus, the goal of this third step is to select the best node for geoparsing among all the candidates returned by the expansion step, as sketched in Figure~\mbox{\ref{fig:gsp-steps-sel}}.

In~\cite{avvenuti2018gsp}, results of the vertical expansion were filtered by a binary SVM classifier. This simple solution was successful because of the limited number of candidates yielded by the vertical expansion. In our present work however, the novel horizontal expansion potentially yields several orders of magnitude more candidates, thus making a binary classification task extremely unbalanced, hence impractical. For this reason, here we cast the selection problem as a regression task, where we aim to predict a confidence score for each candidate node. After assigning a confidence to each candidate returned by the expansion step, \texttt{GSP} simply selects the node with the highest confidence and geotags the input document with the geographic coordinates of that node. This step is conceptually similar to a filtering/pruning step -- also adopted in many machine learning algorithms for improving the accuracy of predictions -- where \texttt{GSP} selects the entity for which it is more confident. Notably, the efficacy of the \textit{expansion\&selection} approach has already been demonstrated in~\cite{avvenuti2018gsp}, with the selection/filtering step significantly boosting the model's \textit{precision}.
In Section~\ref{sec:selection} we further elaborate on this step, discussing how we frame the regression task, how we train our model, and which features we leverage for the regression.

\subsubsection{Concluding steps}
\label{sec:final-notes}
The core of the proposed \texttt{GSP} technique was outlined in the two previous sections. However, \texttt{GSP} performs two additional simple operations before outputting its predictions, described in the following for completeness. The vertical and horizontal expansion strategies are orthogonal with respect to one another, meaning that they give access to different nodes and, potentially, to different information. Being orthogonal, they can also be used simultaneously. Because of this, after selecting a node with the approach described in the previous section, \texttt{GSP} also applies the vertical expansion. In other words, our technique actually leverages both the horizontal and vertical expansions, as shown in Figure~\ref{fig:gsp-expansion}.

The extraction of the geographic coordinates from a node in a knowledge graph, occurs by means of a \textit{parsing} step. In this step, \texttt{GSP} scans all predicates of the semantic resource, looking for geographic information. In particular, in Linked Data there exist many different predicates designed to store geographic information (e.g., \texttt{geo:lat} and \texttt{geo:long}, \texttt{georss:point}, etc.). In \texttt{GSP}, we support as many as 45 geographic predicates. Moreover, since the geographic information contained in such predicates can be stored in different formats (e.g., decimal degrees; degrees, minutes, seconds), we then implemented a set of simple formulas for converting the different geographic formats into \textit{decimal latitude and longitude} coordinates. As a result, the output of the \textit{parsing} step and of the whole \texttt{GSP} technique is represented by a decimal latitude and longitude geographic coordinate (wherever available) that complements the input document. 
 \section{Information expansion strategies}
\label{sec:expansion}
In this section, we describe the different strategies leveraged by \texttt{GSP} as part of its \textit{expansion} step. Each strategy follows a different intuition, with the goal of retrieving the largest set of geographic entities related to the starting one. In the last part of this section we compare the effectiveness of the different strategies, both when applied individually and jointly. To better clarify each strategy and the differences between them, we make use of a fictitious toy experiment. Let us assume to have the following short text to geoparse:
\begin{quote}
    \centering
    \textit{``I'll spend a couple of hours in Bath, visiting its Roman heritage.''}
\end{quote}
The ground truth for this text is represented by the geographic coordinates of the British town of \textit{Bath}, renowned for its thermal baths dating back to the Roman Empire. However, let us assume for this example that the semantic annotator fails in linking the token \textit{Bath} to the correct entity\footnote{\scriptsize\url{http://dbpedia.org/page/Bath,_Somerset}}. Instead, it erroneously links \textit{Bath} to the entity \texttt{bath}\footnote{\scriptsize\url{http://dbpedia.org/page/Bathtub}} (in the sense of bathtub), which is the starting node in our toy example. In the following, we apply the different geographic expansion strategies with size $L=2$ to the small toy knowledge graph repeated once per strategy in Figure~\ref{fig:exp-toy-example}.

\subsection{Spelling-based expansion}
\label{sec:exp-spelling}
Although semantic annotators are specifically designed to disambiguate entities, polysemy, typos and jargon still pose a challenge. The majority of disambiguation errors occurs between entities with very similar names. A natural choice for expanding our set of nodes and for correcting possible errors, is therefore to consider entities whose names are similar to that of the starting node.

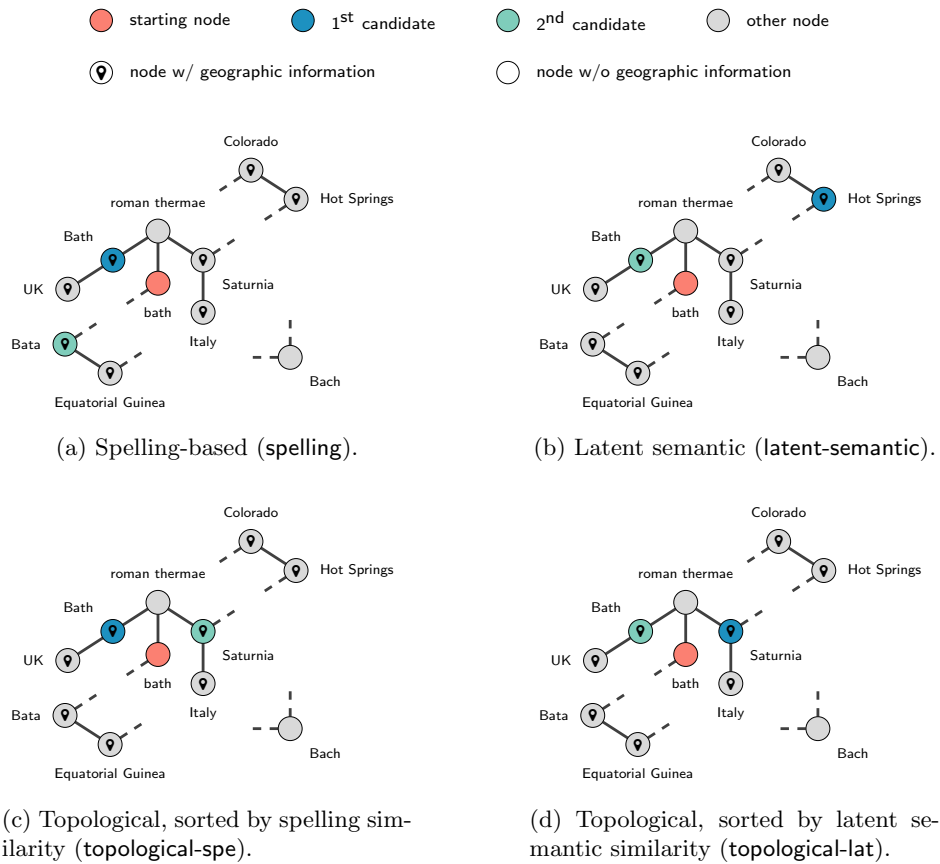
\begin{figure*}[t]
    \centering
\begin{subfigure}[t]{0.65\textwidth}
        \centering
        \captionsetup[subfigure]{labelformat=empty}
        \resizebox{1\textwidth}{!}{\begin{tikzpicture}[
	mycircle/.style={
	circle,
	draw=black,
	text opacity=1,
	inner sep=0pt,
	minimum size=6.5pt}]
	
\node[mycircle,ultra thin,fill=starting] (sta) {};
	\node[right=0.05cm of sta] (lab1) {\tiny starting node};
	
	\node[mycircle,ultra thin,fill=expanded2,right=0.5cm of lab1] (exp1) {};
	\node[right=0.05cm of exp1] (lab2) {\tiny 1\textsuperscript{st} candidate};

	\node[mycircle,ultra thin,fill=expanded4,right=0.5cm of lab2] (exp2) {};
	\node[right=0.05cm of exp2] (lab3) {\tiny 2\textsuperscript{nd} candidate};

	\node[mycircle,ultra thin,fill=other,right=0.5cm of lab3] (oth) {};
	\node[right=0.05cm of oth] (lab4) {\tiny other node};
	
	\node[mycircle,ultra thin,below=0.3cm of sta] (geo) {\tiny\faMapMarker};
	\node[right=0.05cm of geo] (lab5) {\tiny node w/ geographic information};
    
	\node[mycircle,ultra thin,below=0.3cm of exp2] (nogeo) {};
	\node[right=0.05cm of nogeo] (lab6) {\tiny node w/o geographic information};
\end{tikzpicture} }
    \end{subfigure}\\\par\bigskip
\begin{subfigure}[t]{0.35\textwidth}
        \centering
        \resizebox{1\textwidth}{!}{\begin{tikzpicture}[
	mynode/.style={
	circle,
	draw=black,
	text opacity=1,
	inner sep=0pt,
	minimum size=8.5pt}]
	
\node[mynode,ultra thin,fill=starting] (bath) {};
	\node[below=0.025cm of bath] (lab1) {\tiny bath};
\node[mynode,ultra thin,fill=other,above=0.35cm of bath] (rom) {};
	\node[above=0.025cm of rom] (lab2) {\tiny roman thermae};
\node[mynode,ultra thin,fill=expanded2,below left=0.15cm and 0.35cm of rom] (Bath) {\tiny\faMapMarker};
	\node[above left=0.01cm and 0.01cm of Bath] (lab3) {\tiny Bath};
\node[mynode,ultra thin,fill=other,below left=0.15cm and 0.35cm of Bath] (uk) {\tiny\faMapMarker};
	\node[left=0.025cm of uk] (lab4) {\tiny UK};
\node[mynode,ultra thin,fill=other,below right=0.15cm and 0.35cm of rom] (sat) {\tiny\faMapMarker};
	\node[below right=0.01cm and 0.01cm of sat] (lab5) {\tiny Saturnia};
\node[mynode,ultra thin,fill=other,below=0.35cm of sat] (ita) {\tiny\faMapMarker};
	\node[below=0.025cm of ita] (lab6) {\tiny Italy};
	
\node[below left=0.15cm and 0.35cm of bath] (dummy1) {};
	\node[below left=0.15cm and 0.35cm of ita] (dummy2) {};
	\node[above right=0.15cm and 0.35cm of sat] (dummy3) {};
	
\node[mynode,ultra thin,fill=expanded4,below left=0.15cm and 0.35cm of dummy1] (Bata) {\tiny\faMapMarker};
	\node[left=0.025cm of Bata] (lab7) {\tiny Bata};
\node[mynode,ultra thin,fill=other,below right=0.15cm and 0.35cm of Bata] (eqg) {\tiny\faMapMarker};
	\node[below=0.025cm of eqg] (lab8) {\tiny Equatorial Guinea};
	
\node[mynode,ultra thin,fill=other,above right=0.15cm and 0.35cm of dummy3] (hot) {\tiny\faMapMarker};
	\node[right=0.025cm of hot] (lab9) {\tiny Hot Springs};
\node[mynode,ultra thin,fill=other,above left=0.15cm and 0.35cm of hot] (col) {\tiny\faMapMarker};
	\node[above=0.025cm of col] (lab10) {\tiny Colorado};
	
\node[mynode,ultra thin,fill=other,below right=0.35cm and 0.875cm of ita] (bac) {};
	\node[below right=0.01cm and 0.01cm of bac] (lab11) {\tiny Bach};
	
\node[below left=0.15cm and 0.35cm of col] (dummy4) {};
	\node[above=0.35cm of bac] (dummy5) {};
	\node[left=0.35cm of bac] (dummy6) {};
	
\Edge[lw=1pt](bath)(rom)
	\Edge[lw=1pt](rom)(Bath)
	\Edge[lw=1pt](Bath)(uk)
	\Edge[lw=1pt](rom)(sat)
	\Edge[lw=1pt](sat)(ita)
	
	\Edge[lw=1pt](Bata)(eqg)
	\Edge[lw=1pt](hot)(col)
	
\Edge[lw=1pt,style={dashed}](bath)(dummy1)
	\Edge[lw=1pt,style={dashed}](Bata)(dummy1)
	\Edge[lw=1pt,style={dashed}](eqg)(dummy2)
	\Edge[lw=1pt,style={dashed}](sat)(dummy3)
	\Edge[lw=1pt,style={dashed}](hot)(dummy3)
	\Edge[lw=1pt,style={dashed}](col)(dummy4)
	\Edge[lw=1pt,style={dashed}](bac)(dummy5)
	\Edge[lw=1pt,style={dashed}](bac)(dummy6)
\end{tikzpicture} }
        \caption{Spelling-based (\textsf{spelling}).}
        \label{fig:exp-spelling}
    \end{subfigure}\hspace{0.1\textwidth}\begin{subfigure}[t]{0.35\textwidth}
        \centering
        \resizebox{1\textwidth}{!}{\begin{tikzpicture}[
	mynode/.style={
	circle,
	draw=black,
	text opacity=1,
	inner sep=0pt,
	minimum size=8.5pt}]
	
\node[mynode,ultra thin,fill=starting] (bath) {};
	\node[below=0.025cm of bath] (lab1) {\tiny bath};
\node[mynode,ultra thin,fill=other,above=0.35cm of bath] (rom) {};
	\node[above=0.025cm of rom] (lab2) {\tiny roman thermae};
\node[mynode,ultra thin,fill=expanded4,below left=0.15cm and 0.35cm of rom] (Bath) {\tiny\faMapMarker};
	\node[above left=0.01cm and 0.01cm of Bath] (lab3) {\tiny Bath};
\node[mynode,ultra thin,fill=other,below left=0.15cm and 0.35cm of Bath] (uk) {\tiny\faMapMarker};
	\node[left=0.025cm of uk] (lab4) {\tiny UK};
\node[mynode,ultra thin,fill=other,below right=0.15cm and 0.35cm of rom] (sat) {\tiny\faMapMarker};
	\node[below right=0.01cm and 0.01cm of sat] (lab5) {\tiny Saturnia};
\node[mynode,ultra thin,fill=other,below=0.35cm of sat] (ita) {\tiny\faMapMarker};
	\node[below=0.025cm of ita] (lab6) {\tiny Italy};
	
\node[below left=0.15cm and 0.35cm of bath] (dummy1) {};
	\node[below left=0.15cm and 0.35cm of ita] (dummy2) {};
	\node[above right=0.15cm and 0.35cm of sat] (dummy3) {};
	
\node[mynode,ultra thin,fill=other,below left=0.15cm and 0.35cm of dummy1] (Bata) {\tiny\faMapMarker};
	\node[left=0.025cm of Bata] (lab7) {\tiny Bata};
\node[mynode,ultra thin,fill=other,below right=0.15cm and 0.35cm of Bata] (eqg) {\tiny\faMapMarker};
	\node[below=0.025cm of eqg] (lab8) {\tiny Equatorial Guinea};
	
\node[mynode,ultra thin,fill=expanded2,above right=0.15cm and 0.35cm of dummy3] (hot) {\tiny\faMapMarker};
	\node[right=0.025cm of hot] (lab9) {\tiny Hot Springs};
\node[mynode,ultra thin,fill=other,above left=0.15cm and 0.35cm of hot] (col) {\tiny\faMapMarker};
	\node[above=0.025cm of col] (lab10) {\tiny Colorado};
	
\node[mynode,ultra thin,fill=other,below right=0.35cm and 0.875cm of ita] (bac) {};
	\node[below right=0.01cm and 0.01cm of bac] (lab11) {\tiny Bach};
	
\node[below left=0.15cm and 0.35cm of col] (dummy4) {};
	\node[above=0.35cm of bac] (dummy5) {};
	\node[left=0.35cm of bac] (dummy6) {};
	
\Edge[lw=1pt](bath)(rom)
	\Edge[lw=1pt](rom)(Bath)
	\Edge[lw=1pt](Bath)(uk)
	\Edge[lw=1pt](rom)(sat)
	\Edge[lw=1pt](sat)(ita)
	
	\Edge[lw=1pt](Bata)(eqg)
	\Edge[lw=1pt](hot)(col)
	
\Edge[lw=1pt,style={dashed}](bath)(dummy1)
	\Edge[lw=1pt,style={dashed}](Bata)(dummy1)
	\Edge[lw=1pt,style={dashed}](eqg)(dummy2)
	\Edge[lw=1pt,style={dashed}](sat)(dummy3)
	\Edge[lw=1pt,style={dashed}](hot)(dummy3)
	\Edge[lw=1pt,style={dashed}](col)(dummy4)
	\Edge[lw=1pt,style={dashed}](bac)(dummy5)
	\Edge[lw=1pt,style={dashed}](bac)(dummy6)
\end{tikzpicture} }
        \caption{Latent semantic (\textsf{latent-semantic}).}
        \label{fig:exp-latsem}
    \end{subfigure}\par\bigskip
\begin{subfigure}[t]{0.35\textwidth}
        \centering
        \resizebox{1\textwidth}{!}{\begin{tikzpicture}[
	mynode/.style={
	circle,
	draw=black,
	text opacity=1,
	inner sep=0pt,
	minimum size=8.5pt}]
	
\node[mynode,ultra thin,fill=starting] (bath) {};
	\node[below=0.025cm of bath] (lab1) {\tiny bath};
\node[mynode,ultra thin,fill=other,above=0.35cm of bath] (rom) {};
	\node[above=0.025cm of rom] (lab2) {\tiny roman thermae};
\node[mynode,ultra thin,fill=expanded2,below left=0.15cm and 0.35cm of rom] (Bath) {\tiny\faMapMarker};
	\node[above left=0.01cm and 0.01cm of Bath] (lab3) {\tiny Bath};
\node[mynode,ultra thin,fill=other,below left=0.15cm and 0.35cm of Bath] (uk) {\tiny\faMapMarker};
	\node[left=0.025cm of uk] (lab4) {\tiny UK};
\node[mynode,ultra thin,fill=expanded4,below right=0.15cm and 0.35cm of rom] (sat) {\tiny\faMapMarker};
	\node[below right=0.01cm and 0.01cm of sat] (lab5) {\tiny Saturnia};
\node[mynode,ultra thin,fill=other,below=0.35cm of sat] (ita) {\tiny\faMapMarker};
	\node[below=0.025cm of ita] (lab6) {\tiny Italy};
	
\node[below left=0.15cm and 0.35cm of bath] (dummy1) {};
	\node[below left=0.15cm and 0.35cm of ita] (dummy2) {};
	\node[above right=0.15cm and 0.35cm of sat] (dummy3) {};
	
\node[mynode,ultra thin,fill=other,below left=0.15cm and 0.35cm of dummy1] (Bata) {\tiny\faMapMarker};
	\node[left=0.025cm of Bata] (lab7) {\tiny Bata};
\node[mynode,ultra thin,fill=other,below right=0.15cm and 0.35cm of Bata] (eqg) {\tiny\faMapMarker};
	\node[below=0.025cm of eqg] (lab8) {\tiny Equatorial Guinea};
	
\node[mynode,ultra thin,fill=other,above right=0.15cm and 0.35cm of dummy3] (hot) {\tiny\faMapMarker};
	\node[right=0.025cm of hot] (lab9) {\tiny Hot Springs};
\node[mynode,ultra thin,fill=other,above left=0.15cm and 0.35cm of hot] (col) {\tiny\faMapMarker};
	\node[above=0.025cm of col] (lab10) {\tiny Colorado};
	
\node[mynode,ultra thin,fill=other,below right=0.35cm and 0.875cm of ita] (bac) {};
	\node[below right=0.01cm and 0.01cm of bac] (lab11) {\tiny Bach};
	
\node[below left=0.15cm and 0.35cm of col] (dummy4) {};
	\node[above=0.35cm of bac] (dummy5) {};
	\node[left=0.35cm of bac] (dummy6) {};
	
\Edge[lw=1pt](bath)(rom)
	\Edge[lw=1pt](rom)(Bath)
	\Edge[lw=1pt](Bath)(uk)
	\Edge[lw=1pt](rom)(sat)
	\Edge[lw=1pt](sat)(ita)
	
	\Edge[lw=1pt](Bata)(eqg)
	\Edge[lw=1pt](hot)(col)
	
\Edge[lw=1pt,style={dashed}](bath)(dummy1)
	\Edge[lw=1pt,style={dashed}](Bata)(dummy1)
	\Edge[lw=1pt,style={dashed}](eqg)(dummy2)
	\Edge[lw=1pt,style={dashed}](sat)(dummy3)
	\Edge[lw=1pt,style={dashed}](hot)(dummy3)
	\Edge[lw=1pt,style={dashed}](col)(dummy4)
	\Edge[lw=1pt,style={dashed}](bac)(dummy5)
	\Edge[lw=1pt,style={dashed}](bac)(dummy6)
\end{tikzpicture} }
        \caption{Topological, sorted by spelling similarity (\textsf{topological-spe}).}
        \label{fig:exp-topo-spe}
    \end{subfigure}\hspace{0.1\textwidth}\begin{subfigure}[t]{0.35\textwidth}
        \centering
        \resizebox{1\textwidth}{!}{\begin{tikzpicture}[
	mynode/.style={
	circle,
	draw=black,
	text opacity=1,
	inner sep=0pt,
	minimum size=8.5pt}]
	
\node[mynode,ultra thin,fill=starting] (bath) {};
	\node[below=0.025cm of bath] (lab1) {\tiny bath};
\node[mynode,ultra thin,fill=other,above=0.35cm of bath] (rom) {};
	\node[above=0.025cm of rom] (lab2) {\tiny roman thermae};
\node[mynode,ultra thin,fill=expanded4,below left=0.15cm and 0.35cm of rom] (Bath) {\tiny\faMapMarker};
	\node[above left=0.01cm and 0.01cm of Bath] (lab3) {\tiny Bath};
\node[mynode,ultra thin,fill=other,below left=0.15cm and 0.35cm of Bath] (uk) {\tiny\faMapMarker};
	\node[left=0.025cm of uk] (lab4) {\tiny UK};
\node[mynode,ultra thin,fill=expanded2,below right=0.15cm and 0.35cm of rom] (sat) {\tiny\faMapMarker};
	\node[below right=0.01cm and 0.01cm of sat] (lab5) {\tiny Saturnia};
\node[mynode,ultra thin,fill=other,below=0.35cm of sat] (ita) {\tiny\faMapMarker};
	\node[below=0.025cm of ita] (lab6) {\tiny Italy};
	
\node[below left=0.15cm and 0.35cm of bath] (dummy1) {};
	\node[below left=0.15cm and 0.35cm of ita] (dummy2) {};
	\node[above right=0.15cm and 0.35cm of sat] (dummy3) {};
	
\node[mynode,ultra thin,fill=other,below left=0.15cm and 0.35cm of dummy1] (Bata) {\tiny\faMapMarker};
	\node[left=0.025cm of Bata] (lab7) {\tiny Bata};
\node[mynode,ultra thin,fill=other,below right=0.15cm and 0.35cm of Bata] (eqg) {\tiny\faMapMarker};
	\node[below=0.025cm of eqg] (lab8) {\tiny Equatorial Guinea};
	
\node[mynode,ultra thin,fill=other,above right=0.15cm and 0.35cm of dummy3] (hot) {\tiny\faMapMarker};
	\node[right=0.025cm of hot] (lab9) {\tiny Hot Springs};
\node[mynode,ultra thin,fill=other,above left=0.15cm and 0.35cm of hot] (col) {\tiny\faMapMarker};
	\node[above=0.025cm of col] (lab10) {\tiny Colorado};
	
\node[mynode,ultra thin,fill=other,below right=0.35cm and 0.875cm of ita] (bac) {};
	\node[below right=0.01cm and 0.01cm of bac] (lab11) {\tiny Bach};
	
\node[below left=0.15cm and 0.35cm of col] (dummy4) {};
	\node[above=0.35cm of bac] (dummy5) {};
	\node[left=0.35cm of bac] (dummy6) {};
	
\Edge[lw=1pt](bath)(rom)
	\Edge[lw=1pt](rom)(Bath)
	\Edge[lw=1pt](Bath)(uk)
	\Edge[lw=1pt](rom)(sat)
	\Edge[lw=1pt](sat)(ita)
	
	\Edge[lw=1pt](Bata)(eqg)
	\Edge[lw=1pt](hot)(col)
	
\Edge[lw=1pt,style={dashed}](bath)(dummy1)
	\Edge[lw=1pt,style={dashed}](Bata)(dummy1)
	\Edge[lw=1pt,style={dashed}](eqg)(dummy2)
	\Edge[lw=1pt,style={dashed}](sat)(dummy3)
	\Edge[lw=1pt,style={dashed}](hot)(dummy3)
	\Edge[lw=1pt,style={dashed}](col)(dummy4)
	\Edge[lw=1pt,style={dashed}](bac)(dummy5)
	\Edge[lw=1pt,style={dashed}](bac)(dummy6)
\end{tikzpicture} }
        \caption{Topological, sorted by latent semantic similarity (\textsf{topological-lat}).}
        \label{fig:exp-topo-lat}
    \end{subfigure}\caption{Toy example showing the nodes retrieved by the different expansion strategies on a small knowledge graph, for expansion size $L=2$.} \label{fig:exp-toy-example}
\end{figure*}

For any given starting node, the spelling-based expansion (henceforth \textsf{spelling}) retrieves and sorts the top-$L$ geographic entities having closest names. The similarity between entities names is computed as the case-sensitive Levenshtein (edit) distance. Figure~\ref{fig:exp-spelling} shows the results of \textsf{spelling} when applied to our toy experiment. After sorting the knowledge graph entities, \textsf{spelling} yields the sorted vector $[\texttt{Bath}, \texttt{Bach}, \texttt{Bata}, \dots]$. The entity \texttt{Bach}, corresponding to the famous musician, is not complemented with geographic information and so it is discarded. Finally, \textsf{spelling}$\rvert_{L=2}$ returns the geographic entities corresponding to the British town \texttt{Bath} and to the Equatorial Guinea city \texttt{Bata}. By traversing the graph based on entities names, this strategy can retrieve nodes that are not directly linked nor topologically near to the starting one.

\subsection{Latent semantic expansion}
\label{sec:exp-rdf2vec}
Node embeddings refer to a set of techniques for unsupervised feature extraction from large graphs, inspired by the usefulness that word and document embeddings recently demonstrated for many text mining tasks~\cite{devlin2019bert}. In this representation, each node in a graph is described by a high-dimensional feature vector capable of encoding the latent structural information of the node within the graph. As such, nodes that play a similar role in the topology of a graph end up having similar node embeddings representations. Similarly to word embeddings, several different techniques have been proposed for computing the node embeddings of a graph. Among these, the \textit{rdf2vec} technique~\cite{ristoski2019rdf2vec} was recently proposed and specifically designed for semantic knowledge graphs, such as the ones leveraged by \texttt{GSP}. In particular, \textit{rdf2vec} extends previous generic node embeddings techniques by also considering semantic node properties and the many different types of edges that represent the semantic relations between entities in knowledge graphs. As such, it is particularly suitable for graph mining tasks on knowledge graphs. In our experiments, we used the readily-available \textit{rdf2vec} embeddings\footnote{\scriptsize\url{http://data.dws.informatik.uni-mannheim.de/rdf2vec/}}, pre-trained for the English \textit{DBpedia}.

Latent semantic expansion (henceforth \textsf{latent-semantic}) retrieves and sorts the top-$L$ nodes having largest cosine similarity between their \textit{rdf2vec} representation and that of the starting node. In other words, this expansion strategy leverages powerful semantic node embeddings techniques to retrieve the nodes that are most similar to the starting one, in the latent semantic vector space. Similarly to the \textsf{spelling} expansion, also \textsf{latent-semantic} potentially retrieves nodes that are topologically far from the starting one. When applied to our toy example, we imagine \textsf{latent-semantic}$\rvert_{L=2}$ to retrieve geographic entities that are semantically related to the concept of bathing (e.g., having hot springs), such as 
$[\texttt{Hot Springs}, \texttt{Bath}, \dots]$, as shown in Figure~\ref{fig:exp-latsem}.

\subsection{Topological expansion}
\label{sec:exp-topol}
The previously introduced expansions do not explicitly consider the topology of the knowledge graph. However, when the semantic annotator fails to point to the correct starting node, or when the starting node is not complemented by geographic information, relevant geographic information is nonetheless likely to be found in a topologically near node, with respect to the starting one. In other words, we are confident that the annotator pointed us at least in the vicinity of the correct node. Thus, the following expansion strategies leverage this hypothesis and traverse existing links between nodes.

In practice, to implement topological expansion of size $L$, we retrieve the $L$ nearest nodes with respect to the starting one. We begin by retrieving 1-hop geographic nearest neighbors to the starting node, then 2-hops geographic neighbors and so on, until we retrieve $L$ nodes. Differently from previous strategies, all $n$-hop neighbors of the starting node share the same ``similarity'', which requires a mean to break ties since we want our expansion strategies to yield sorted vectors. To sort nodes that have the same distance with respect to the starting one, we leverage the similarity measures used in the \textsf{spelling} and \textsf{latent-semantic} strategies. In detail, we have the (i) \textsf{topological-spe} strategy, when the sorting criterion is based on spelling (i.e., the edit distance between entities names), and the (ii) \textsf{topological-lat} strategy, when the sorting criterion is based on latent semantic similarity (i.e., the cosine similarity between \textit{rdf2vec} vectors). Interestingly, these two strategies can be seen as the combination of two orthogonal sorting criteria -- namely, \textit{topology} and either \textit{spelling} or \textit{latent semantic} similarity. Figures~\ref{fig:exp-topo-spe} and~\ref{fig:exp-topo-lat} show the results of our topological expansions, when applied to the toy example.

\subsection{Evaluation}
\label{sec:exp-eval}

\begin{figure*}[t]
    \centering
    \begin{subfigure}[t]{0.37\textwidth}
        \centering
        \includegraphics[width=1\textwidth]{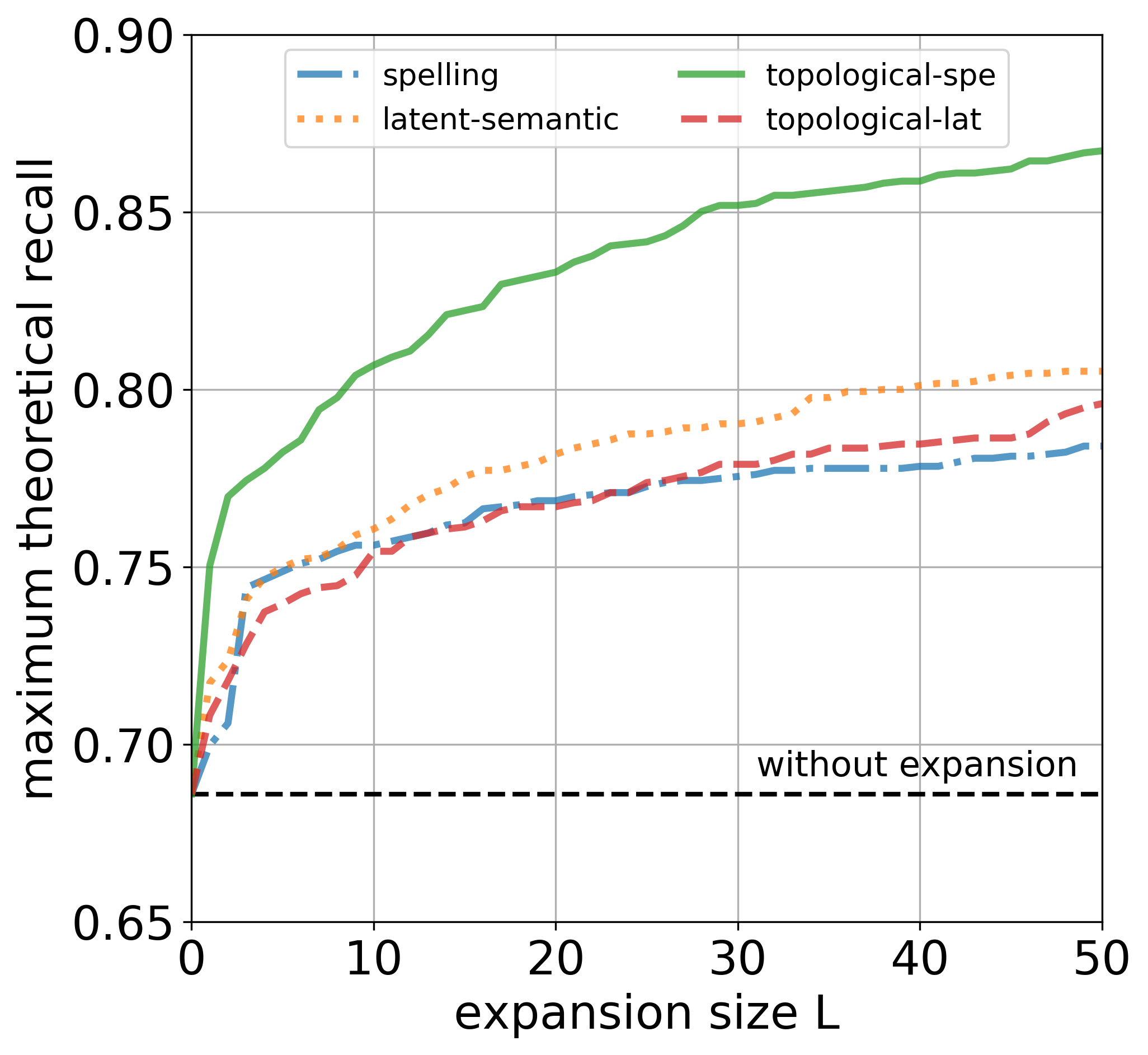}
        \caption{Maximum theoretical recall produced by the different expansion strategies, when applied individually.}
\label{fig:exp-results-individual}
    \end{subfigure}\hspace{0.1\textwidth}\begin{subfigure}[t]{0.37\textwidth}
        \centering
        \includegraphics[width=1\textwidth]{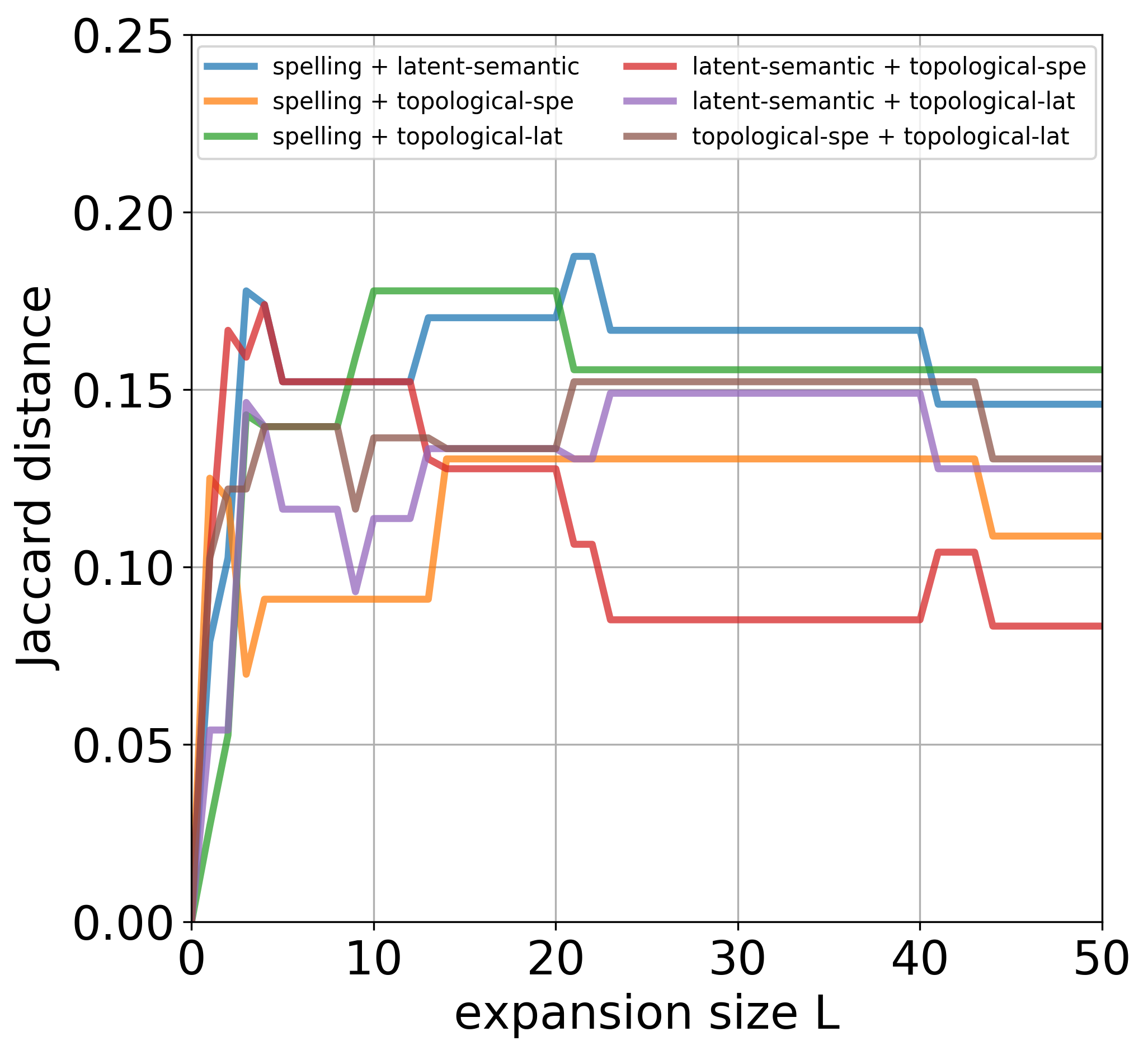}
        \caption{Combined effectiveness of the expansion strategies, measured as the Jaccard distance $d_J$ between the sets of retrieved entities.}
\label{fig:exp-results-joint}
    \end{subfigure}\caption{Performance evaluation of the proposed expansion strategies, when applied individually and jointly, as a function of the expansion size $L$.}
    \label{fig:exp-results}
\end{figure*}

In the remainder of this section, we evaluate the effectiveness of the different expansion strategies by measuring the extent to which they increase the number of geotagged toponyms (i.e, toponyms for which we retrieved geographic coordinates). 

\paragraph{Experimental setup and evaluation metrics} As anticipated in Section~\ref{sec:method-expansion}, increasing geotagged toponyms contributes to raise geoparsing \textit{recall}. Because of this, we 
introduce an \emph{ad hoc} metric to evaluate the expansion strategies, called \textit{maximum theoretical recall}, defined as the recall we would obtain by always selecting the correct geographic coordinates among all those retrieved by a given expansion strategy. In practice, the real recall also depends on the selection step, that might erroneously discard correct geographic information retrieved via an expansion strategy. We evaluate this facet in the next section, while here we only focus on measuring geographic information retrieved via expansion, that would not have been collected otherwise. The effectiveness of the various strategies is evaluated as a function of the expansion size $L$. By definition, the maximum theoretical recall is a monotonic non-decreasing function of $L$. Strategies are compared between one another, and with the baseline scenario where no expansion is performed. In Figure~\ref{fig:exp-results}, the latter scenario corresponds to expansions of size $L=0$. Instead, we remind that expansions of size $L=1$, despite returning only 1 entity as in the no-expansion scenario (where only the starting node is considered), produce improved results since expanding forces to select geographic entities, independently of $L$. The following results are obtained on the training split of our dataset, thoroughly described in Section~\mbox{\ref{sec:res-dataset}}.

\paragraph{Results} Figure~\ref{fig:exp-results-individual} shows the results of the different expansion strategies, when applied individually. Topological expansion with spelling-based sorting (\textsf{topological-spe}) largely outperforms all other strategies. With respect to the baseline scenario where no expansion is applied, \textsf{topological-spe} produces most of the recall gain within the first expansion steps: $+9.4$\% at $L=1$ and $+12.0$\% at $L=2$. This means that, assuming a flawless selection step, the sole application of \textsf{topological-spe}$\rvert_{L=2}$ would boost the overall geoparsing recall by $12$\%. This important finding proves our hypotheses correct and motivates our experimentation on expansion strategies. Largely boosting recall at small expansion sizes is a desirable feature, since the complexity of the subsequent selection task dramatically increases with $L$. For $L\geq3$ the incremental gain at each step is $<1$\%. Nevertheless, \textsf{topological-spe} keeps retrieving relevant geographic information also for larger values of $L$, as demonstrated by its steadily rising curve in Figure~\ref{fig:exp-results-individual}, up to a recall gain of $+26.3$\% at $L=50$. All other expansion strategies achieve significantly worse results, demonstrating a lower capacity of retrieving relevant geographic information from their exploration of the knowledge graph. The \textsf{latent-semantic} strategy, based on \textit{rdf2vec} node embeddings, achieves slightly better results with respect to the remaining strategies for $L>10$. Notably, even the worst-performing strategy (i.e., \textsf{spelling}) yields a recall improvement of $+2.0$\% at $L=1$ and $+2.9$\% at $L=2$, further supporting the usefulness of this approach.

\paragraph{Complementarity of expansion strategies} The proposed strategies are diverse and largely orthogonal. As such, they potentially return very different sets of entities for the same input. We are thus interested in evaluating the extent to which two different strategies can complement each other by retrieving complementary information. In other words, two strategies could be individually weak, but when used simultaneously, they could nonetheless provide large amounts of relevant information. To evaluate this facet, we consider the output of all expansion strategies as sets (instead of ordered vectors) of entities. Then, for each combination of two different strategies, we compute the Jaccard distance $d_J$ between the sets of retrieved entities. The higher is $d_J$, which is defined in the $[0, 1]$ range, the more diverse are the entities retrieved by the two strategies. In turn, a large $d_J$ would support their combined application. Figure~\ref{fig:exp-results-joint} shows results of this experiment, as a function of the expansion size $L$. As shown, no combination of strategies achieves a large Jaccard distance, with all combinations laying in the region of $d_J<0.2$. The best results are achieved by \textsf{spelling}$+$\textsf{topological-lat} and by \textsf{spelling}$+$\textsf{latent-semantic}. However, the overall results of this experiment suggest that na\"{i}vely combining different strategies does not produce outright better results.

Given the results of the experiments with the different expansion strategies, in the remainder of our work we perform the expansion step of \texttt{GSP} by applying the topological expansion with spelling-based sorting (\textsf{topological-spe}), which achieved remarkable results also for low $L$ values.

 \section{Best candidate selection}
\label{sec:selection}

\begin{figure*}[t]
    \centering
\begin{subfigure}[t]{0.13\textwidth}
        \centering
        \captionsetup[subfigure]{labelformat=empty}
        \resizebox{1\textwidth}{!}{\begin{tikzpicture}[
	mynode/.style={
	circle,
	draw=black,
	text opacity=1,
	inner sep=0pt,
	minimum size=8.5pt,
	on grid},
	mycenter/.style={
	circle,
	draw=black,
	fill=starting,
	inner sep=0pt,
	minimum size=4pt,
	on grid}]
	
\node[mycenter,ultra thin] (center) {};
	\node[below=0.05cm of center] (lab1) {\tiny ground truth};
    
\node[mynode,ultra thin,fill=expanded1,below left=0.90cm and 0.40cm of center] (node1) {};
	\node[mynode,ultra thin,fill=expanded2,below left=0.90cm and 0.20cm of center] (node2) {};
	\node[mynode,ultra thin,fill=expanded3,below=0.90cm of center] (node3) {};
	\node[mynode,ultra thin,fill=expanded4,below right=0.90cm and 0.20cm of center] (node4) {};
	\node[mynode,ultra thin,fill=expanded5,below right=0.90cm and 0.40cm of center] (node5) {};
	\node[below=0.05cm of node3] (lab2) {\tiny geo entities retrieved};
	\node[below=-0.03cm of lab2,inner sep=0pt] (lab2bis) {\tiny via expansion};
\end{tikzpicture} }
    \end{subfigure}\hspace{0.07\textwidth}\begin{subfigure}[t]{0.24\textwidth}
        \centering
        \captionsetup[subfigure]{labelformat=empty}
        \resizebox{1\textwidth}{!}{\begin{tikzpicture}[
	mynode/.style={
	circle,
	draw=black,
	text opacity=1,
	inner sep=0pt,
	minimum size=8.5pt,
	on grid},
	mycenter/.style={
	circle,
	draw=black,
	fill=starting,
	inner sep=0pt,
	minimum size=4pt,
	on grid}]
	
\node[mycenter,ultra thin] (center) {};
\draw[black,dashed,postaction={decorate, decoration={text along path,
        text={|\tiny|true positive distance threshold {$\mathcal{T}$}}, text align=center, reverse path=true,raise=0.5ex}}] (0,0) circle[radius=41pt];
    
\node[mynode,ultra thin,fill=expanded1,above left=0.30cm and 0.70cm of center] (node1) {};
	\node[below=0.1cm of node1,inner sep=0pt] (lab1) {\tiny $c=L$};
	\node[mynode,ultra thin,fill=expanded4,above left=1.10cm and 0.30cm of center] (node2) {};
	\node[right=0.1cm of node2,inner sep=1pt,fill=white] (lab2) {\tiny $c=L-2$};
	\node[mynode,ultra thin,fill=expanded2,above right=0.40cm and 0.90cm of center] (node3) {};
	\node[right=0.1cm of node3,inner sep=1pt,fill=white] (lab3) {\tiny $c=L-1$};
	\node[mynode,ultra thin,fill=expanded5,below right=1.60cm and 0.30cm of center] (node4) {};
	\node[below=0.1cm of node4,inner sep=0pt] (lab4) {\tiny $c=0$};
	\node[mynode,ultra thin,fill=expanded3,below right=1.90cm and 1.50cm of center] (node5) {};
	\node[below=0.1cm of node5,inner sep=0pt] (lab5) {\tiny $c=0$};
	
\Edge[Direct,lw=1pt](center)(node1)
	\Edge[Direct,lw=1pt](center)(node2)
	\Edge[Direct,lw=1pt](center)(node3)
	\Edge[Direct,lw=1pt,Math,label=d_g,fontscale=.85](center)(node4)
	\Edge[Direct,lw=1pt](center)(node5)
\end{tikzpicture} }
    \end{subfigure}\caption{Assignment of regression labels to candidates retrieved by an expansion strategy of size $L=5$.}\label{fig:sel-labeling}
\end{figure*}
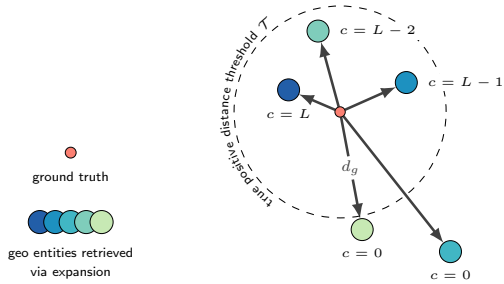

Given a set of $L$ candidate geographic entities retrieved by an expansion strategy, the goal of the \textit{selection} step is to choose the best candidate for geotagging the input document. Based on our problem definition, good candidates are those whose geographic distance $d_g$ from the ground truth coordinates is $<\mathcal{T}$. In fact, all candidates that verify this assumption yield a \textit{true positive} prediction at evaluation time. Among these, the best candidate is arguably the one with the lowest geographic distance from the ground truth.

\subsection{Candidates labeling}
\label{sec:sel-labeling}
As anticipated, we cast the selection problem as a regression task where we estimate a confidence score $\hat{c}$ for each candidate, and we choose the candidate with the highest score. For any given candidate retrieved by an expansion strategy, its ground truth confidence score $c$ should reflect its quality with respect to the ground truth coordinates. In this way, a regression model trained to estimate $c$ scores would in fact produce a geographic ranking of the candidates. We leverage these observations by assigning a ground truth confidence score $c=0$ to candidates whose distance from the ground truth coordinates is $d_g\geq\mathcal{T}$, as shown in the example of Figure~\ref{fig:sel-labeling}. Instead, all candidates whose $d_g<\mathcal{T}$ are assigned a positive score. In particular, the nearest candidate to the ground truth coordinates has $c=L$, the second nearest has $c=L-1$, and so on.

\subsection{Feature engineering}
\label{sec:sel-features}

\begin{table}[!ht]
  \scriptsize
  \centering
  \begin{tabular}{lrlr}
    \toprule
    \multirow{4}{*}{\STAB{\rotatebox[origin=c]{90}{\texttt{A\&E}}}}
     & 1 & \multicolumn{2}{l}{\textsf{confidence:} semantic annotator confidence score} \\
     & 2 & \multicolumn{2}{l}{\textsf{hop:} topological distance between the starting node and the candidate} \\
     & 3 & \multicolumn{2}{l}{\textsf{expansion\_rank:} number of entities traversed by the expansion strategy to reach the candidate} \\
     & 4 & \multicolumn{2}{l}{\textsf{expansion\_rank\_onlygeo:} number of geographic entities traversed by the expansion strategy to reach the candidate} \\
     \midrule
    \multirow{10}{*}{\STAB{\rotatebox[origin=c]{90}{\texttt{SPE}}}}
     & 5 & \multicolumn{2}{l}{\textsf{num\_tokens\_candidate\_label:} number of tokens in the candidate entity name} \\
     & 6 & \multicolumn{2}{l}{\textsf{len\_candidate\_label:} number of characters in the candidate entity name} \\
     & 7 & \multicolumn{2}{l}{\textsf{edit\_from\_original\_label:} edit distance between the names of the starting node and of the candidate} \\
     & 8 & \multicolumn{2}{l}{\textsf{num\_tokens\_anchor:} number of tokens in the anchor} \\
     & 9 & \multicolumn{2}{l}{\textsf{len\_anchor:} number of characters in the anchor} \\
     & 10 & \multicolumn{2}{l}{\textsf{uppercase\_in\_anchor:} number of uppercase characters in the anchor} \\
     & 11 & \multicolumn{2}{l}{\textsf{edit\_from\_anchor:} edit distance between the candidate entity name and the anchor} \\
     & 12 & \multicolumn{2}{l}{\textsf{edit\_ratio\_from\_anchor:} ratio between \textsf{edit\_from\_anchor} and \textsf{len\_anchor}} \\
     & 13 & \multicolumn{2}{l}{\textsf{num\_tokens\_ratio:} ratio between \textsf{num\_tokens\_candidate\_label} and \textsf{num\_tokens\_anchor}} \\
     & 14 & \multicolumn{2}{l}{\textsf{len\_ratio:} ratio between \textsf{len\_candidate\_label} and \textsf{len\_anchor}} \\
    \midrule
    \multirow{9}{*}{\STAB{\rotatebox[origin=c]{90}{\texttt{DBP}}}}
     & 15 & \textsf{superclass:} \textit{DBpedia} ontology class of the candidate entity, derived from \texttt{owl:Thing} subclasses & \textsf{[categorical, 5]} \\
     & 16 & \multicolumn{2}{l}{\textsf{num\_of\_superclasses:} number of superclasses of the \textit{DBpedia} candidate entity} \\
     & 17 & \multicolumn{2}{l}{\textsf{num\_of\_classes:} number of classes of the \textit{DBpedia} candidate entity} \\
     & 18 & \multicolumn{2}{l}{\textsf{page\_degree:} node degree of the \textit{DBpedia} candidate entity} \\
     & 19 & \multicolumn{2}{l}{\textsf{page\_length:} number of characters contained in the corresponding \textit{Wikipedia} article source} \\
     & 20 & \multicolumn{2}{l}{\textsf{anchor\_in\_short\_abstract:} num. occurrences of the anchor in the short abstract of the candidate entity} \\
     & 21 & \textsf{anchor\_in\_short\_abstract\_ci:} num. occurrences of the anchor in the short abstract of the candidate entity & \textsf{[case insensitive]} \\
     & 22 & \multicolumn{2}{l}{\textsf{anchor\_in\_long\_abstract:} num. occurrences of the anchor in the long abstract of the candidate entity} \\
     & 23 & \textsf{anchor\_in\_long\_abstract\_ci:} num. occurrences of the anchor in the long abstract of the candidate entity & \textsf{[case insensitive]} \\
    \midrule
    \multirow{4}{*}{\STAB{\rotatebox[origin=c]{90}{\texttt{SYN}}}}
     & 24 & \textsf{pos\_tag:} part-of-speech (POS) tag of the anchor & \textsf{[categorical, 50]} \\
     & 25 & \textsf{chunk\_tag:} chunking tag of the anchor & \textsf{[categorical, 10]} \\
     & 26 & \multicolumn{2}{l}{\textsf{pos\_confidence:} POS-tagging confidence} \\
     & 27 & \multicolumn{2}{l}{\textsf{chunk\_confidence:} chunking confidence} \\
     \midrule
     \multirow{2}{*}{\STAB{\rotatebox[origin=c]{90}{\texttt{NER}}}}
     & 28 & \textsf{ner\_tag:} named-entity recognition (NER) tag of the anchor & \textsf{[categorical, 5]} \\
     & 29 & \multicolumn{2}{l}{\textsf{ner\_confidence:} NER-tagging confidence} \\
     \midrule
     \multirow{2}{*}{\STAB{\rotatebox[origin=c]{90}{\texttt{LAT}}}}
     & 30 & \multicolumn{2}{l}{\textsf{rdf2vec\_similarity:} cosine similarity between the starting node's and the candidate's \textit{rdf2vec} vectors} \\
     & 31 & \multicolumn{2}{l}{\textsf{bert\_similarity:} cosine similarity between the anchor's and the candidate's name BERT vectors} \\
     \bottomrule
  \end{tabular}
  \caption{Grouping and brief description of the features used for the regression task. We specify categorical features, together with their cardinality, and case insensitive features.}
\label{tab:sel-features}
\end{table}  
In the previous section we described how ground truth labels for the regression task are assigned. Now, we list and describe the features for our regression model (i.e., our regressors). Table~\ref{tab:sel-features} lists and briefly describes the features that we compute for each candidate. Each feature aims at measuring the relations between the candidate and (i) the starting node given by the semantic annotator, (ii) the token(s) of the input document that were linked to the starting node by the annotator (henceforth called the \textit{anchor}), and (iii) the overall context (including the anchor's context in the input document and the nodes context in the knowledge graph). Notably, our features do not leverage any geographic information, implying that we aim to estimate the geographic quality of candidates based on other characteristics (i.e., their relations with the starting node, the anchor and their context). In the following and in Table~\ref{tab:sel-features}, we group features according to the type of information they convey.

\paragraph{Annotation \& expansion features \textup{(\texttt{A\&E})}} These features are based on the confidence with which the semantic annotator linked the anchor to the starting node in the knowledge graph, and on the distance traversed by the expansion strategy to retrieve the candidate. A large annotation confidence usually implies that the starting node already represents a good candidate for geoparsing. Conversely, a poor confidence is a proxy for disambiguation errors. In that case, considering other entities may be advantageous. Similarly, the further the expansion moves away from the starting node, the less the retrieved candidate is semantically related to it. As a consequence, large distances may discriminate non pertinent resources.

\paragraph{Spelling features \textup{(\texttt{SPE})}} Spelling features are designed to capture the spelling characteristics of the anchors and of the entity names, both separately and compared. The rationale behind this class of features is that it is unlikely that the name of a good candidate is completely unrelated to the anchor. At the same time, a high similarity between the candidate and the starting node reflects a possible disambiguation error of the annotator. Finally, location mentions often start with uppercase letters. Hence, we consider their presence in the anchor as possible proxies of location mentions. All these features can be computed solely from the anchor string and the candidate URL, both of which are returned by the semantic annotator.

\paragraph{DBpedia features \textup{(\texttt{DBP})}} This class of features considers the content and the semantic characteristics of the candidate in the reference knowledge graph (i.e., the English \textit{DBpedia}). In particular, we leverage the \textit{DBpedia} ontology to understand the resource type of the candidate (e.g., place, event, activity, agent), with a specific focus on places. Moreover, we measure the centrality of the corresponding node within the graph. Finally, we estimate how much its descriptive abstracts are related to the anchor in the input document.

\paragraph{Syntactic features \textup{(\texttt{SYN})}} We leverage natural language processing (NLP) techniques to carry out a syntactic analysis of the input text, assigning tags to the anchors according to their syntactic role 
(e.g., nouns, verbs, 
\emph{etc.}). In fact, valid location anchors are more likely to correspond to specific syntactic tags such as nouns, with respect to other tags (e.g., verbs, conjunctions). In detail, we perform two types of syntactic analysis: (i) part-of-speech (POS) tagging and (ii) text chunking. POS tagging analyses the text word by word, 
considering a fine-grained set of 50 possible tags. Instead, text chunking identifies and tags syntactically correlated groups of words, with a coarser-grained set of 10 possible tags.
We obtain POS and chunking tags, together with their respective confidence scores, by adopting the state-of-the-art NLP framework FLAIR\footnote{\scriptsize\url{https://github.com/flairNLP/flair}}~\cite{akbik2019flair}.

\paragraph{Named Entity features \textup{(\texttt{NER})}} Named Entity Recognition (NER) is a task that aims to locate NE mentions in unstructured text and to tag them with predefined categories (e.g., persons, organizations, locations). Similarly to syntactic features, we follow the idea that NER tags corresponding to the category \textit{location} are good proxies for valid anchors, whereas anchors tagged differently should be discarded for geoparsing tasks. Again, we leverage FLAIR to perform the NER task on our input texts, obtaining NER tags with the respective confidence. Notably, many state-of-the-art geoparsing techniques are based on NER to identify toponyms in texts~\cite{middleton2014b,halterman2017mordecai,middleton2018location}.

\paragraph{Latent features \textup{(\texttt{LAT})}} The last group of features leverages state-of-the-art embeddings techniques designed for texts and graphs, in order to estimate the latent similarity between the candidate and, respectively, the anchor and the starting node. In Section~\ref{sec:exp-rdf2vec} we introduced \textit{rdf2vec} node embeddings~\cite{ristoski2019rdf2vec} as a profitable mean to retrieve and sort candidate entities for expansion. Here, we leverage the same idea for estimating the similarity between the candidate and the starting node. To do so, we simply compute the cosine similarity between the respective \textit{rdf2vec} vectors. 
Then, in order to assess the semantic similarity between the candidate and the anchor, we leverage contextual word embeddings, and in particular their state-of-the-art implementation BERT\footnote{\scriptsize\url{https://github.com/google-research/bert}}~\cite{devlin2019bert}. Word embeddings are dense, continuous representations of words, designed to capture their distributional and semantic properties. While traditional approaches assign a unique vector to each word, \textit{contextual} word embeddings address the problem of polysemy by computing word vectors depending on the specific context in which words appear. Here, we are interested in assessing how much the concept addressed by the anchor in the input text is related to the candidate entity. Hence, we apply BERT to the input text, extracting the embeddings vector corresponding to the anchor. This vector captures the semantic properties of the anchor with respect to the specific context in which it appears. Similarly, we also apply BERT to the candidate's short abstract, extracting the vector corresponding to the first occurrence of the resource name in the abstract. We obtain an embeddings vector representing the candidate within its abstract. Finally, we compute the cosine similarity between the two aforementioned vectors, thus assessing the semantic similarity between the candidate and the anchor. This feature can help our regression model to address polysemy, by providing different representations of the same tokens appearing in different contexts and by estimating latent semantic similarities between candidates and their corresponding anchors.

\subsection{Regression algorithms}
\label{sec:sel-algorithms}
To train our regression models we resort to state-of-the-art machine learning algorithms based on decision trees. In detail, we experiment with the 3 following algorithms:
\begin{enumerate*}[label=(\roman*)]
    \item Random Forest (RF),
    \item Gradient Boosting Decision Trees (GBDT), and
    \item Dropouts meet multiple Additive Regression Trees (DART).
\end{enumerate*}
RF is a well-known ensemble learning technique based on multiple decision trees. The different trees are trained in parallel, each one receiving as input a random sample of the training instances and of the available features, implementing the so-called \textit{bagging} approach. The outputs of the trees are aggregated by a suitable ensemble method. The bagging approach is able to mitigate the model variance and possible overfitting, making RF an efficient and accurate technique.
GBDT implements the ensemble learning paradigm in a completely different way. During the training phase, GBDT grows a sequence of \textit{weak} learners (i.e., shallow trees), in which each weak learner focuses on correcting the residual errors of the current model approximation. By aggregating the weak learner outputs, GBDT generates a \emph{strong} learner. GBDT often outperforms RF in terms of prediction performance, but it is more computationally expensive and more prone to overfitting.
The last algorithm with which we experiment is DART. It modifies an ensemble learning approach similar to GBDT by introducing \textit{dropout}, a feature borrowed from deep learning. In this context, dropout consists in randomly dropping trees. This strategy proves useful to prevent trivial trees and to mitigate overfitting, but it has a negative impact on computational efficiency~\cite{rashmi2015dart}.

We implemented all the aforementioned algorithms
with the Microsoft's LightGBM\footnote{\scriptsize\url{https://github.com/microsoft/LightGBM}} framework, 
representing the current state-of-the-art for tree-based classification, regression and ranking~\cite{ke2017lightgbm}.

\subsection{Evaluation}
\label{sec:sel-eval}
\paragraph{Experimental setup}
We train our regression models using the training set described in Section~\mbox{\ref{sec:res-dataset}}. For each algorithm, we adopt Randomized Search Cross Validation (RSCV) to explore the hyper-parameters space and optimize algorithms settings. We resort to RSCV since it is more efficient than grid search or manual search, especially in the presence of hyper-parameters sampled from continuous distributions~\cite{bergstra2012random}, as in our case. Tree-based algorithms are also known for variable, non-deterministic behaviors, resulting in a certain amount of variance in their performance across different runs. To account for this behavior, we repeat each experiment 10 times, reporting mean and standard deviation of the results, to gain a more reliable estimation of their performance.

At the beginning of this section we explained that the best candidate is simply chosen as the one with the highest estimated confidence score $\hat{c}$. However, we enforce an additional constraint in order to provide more accurate results. In particular, we discard those candidates whose score $\hat{c} < c_{th}$. The value $c_{th}$ represents a confidence threshold that allows to prune predictions generated with a very low confidence. We calibrate $c_{th}$ to the value that maximizes the model's \textit{F1} on the validation set, specifically created for this purpose. Finally, the obtained models are evaluated on the test set with a fully blind approach. Messages $t_{i}$ belonging to validation and test sets are never used to train the regression models, thus mitigating the risk of overfitting when we calibrate the confidence threshold $c_{th}$, and when we evaluate the model performance.

\paragraph{Evaluation metrics}
We evaluate each selection model as a combination of the related regression and filtering steps. Since the selection concludes the \texttt{GSP}'s analysis pipeline, the most natural choice is to evaluate it according to the same metrics used for the overall geoparsing task. Hence, we adopt the \textit{precision}, \textit{recall} and \textit{F1} metrics, as defined in Section~\mbox{\ref{sec:problem}}.

\begin{figure*}[t]
    \centering
    \begin{subfigure}[t]{0.27\textwidth}
        \centering
        \includegraphics[width=1\textwidth, angle=0]{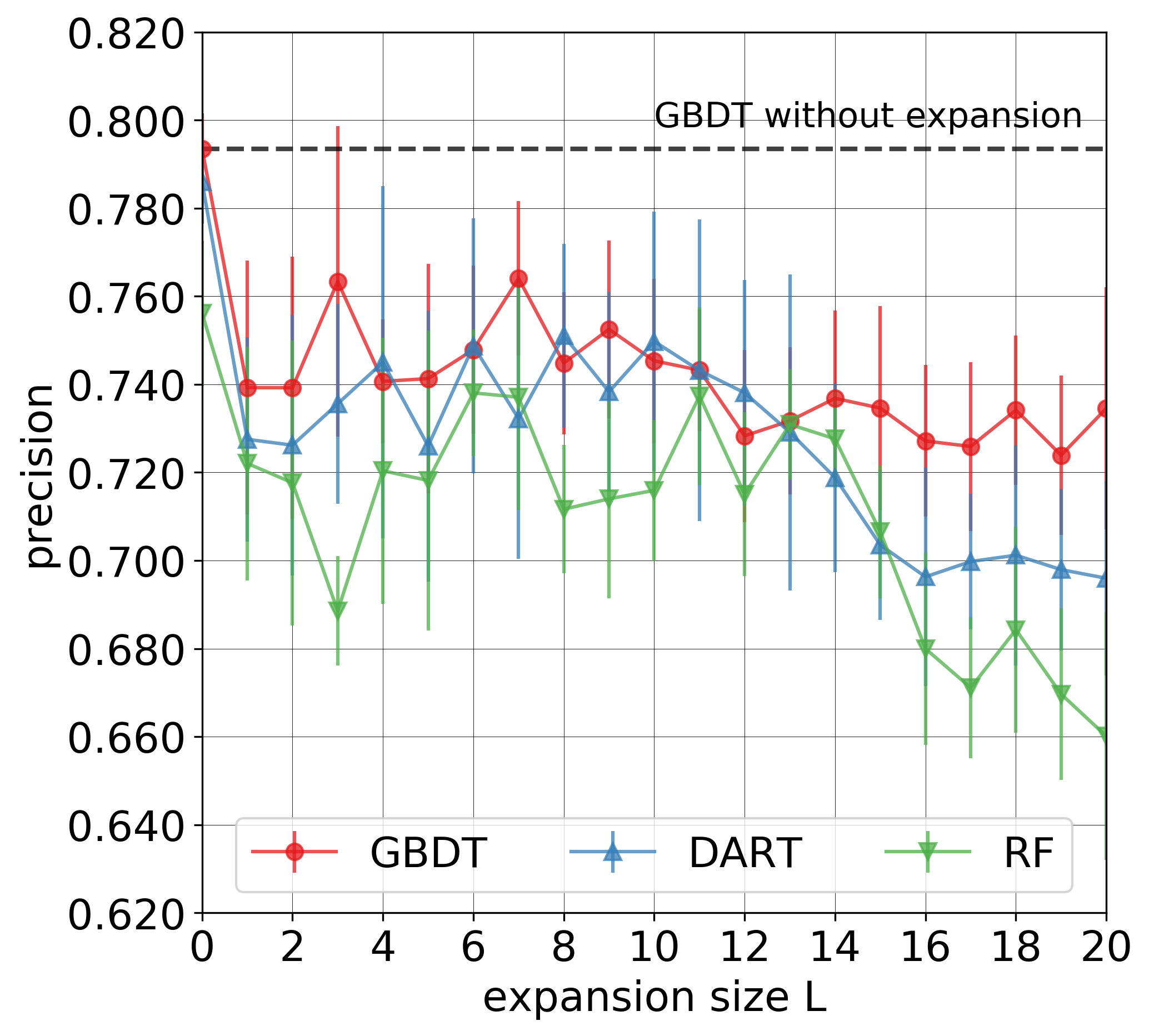}
        \caption{\textit{Precision}.}
        \label{fig:sel-precision}
    \end{subfigure}\hspace{0.04\textwidth}\begin{subfigure}[t]{0.27\textwidth}
        \centering
        \includegraphics[width=1\textwidth, angle=0]{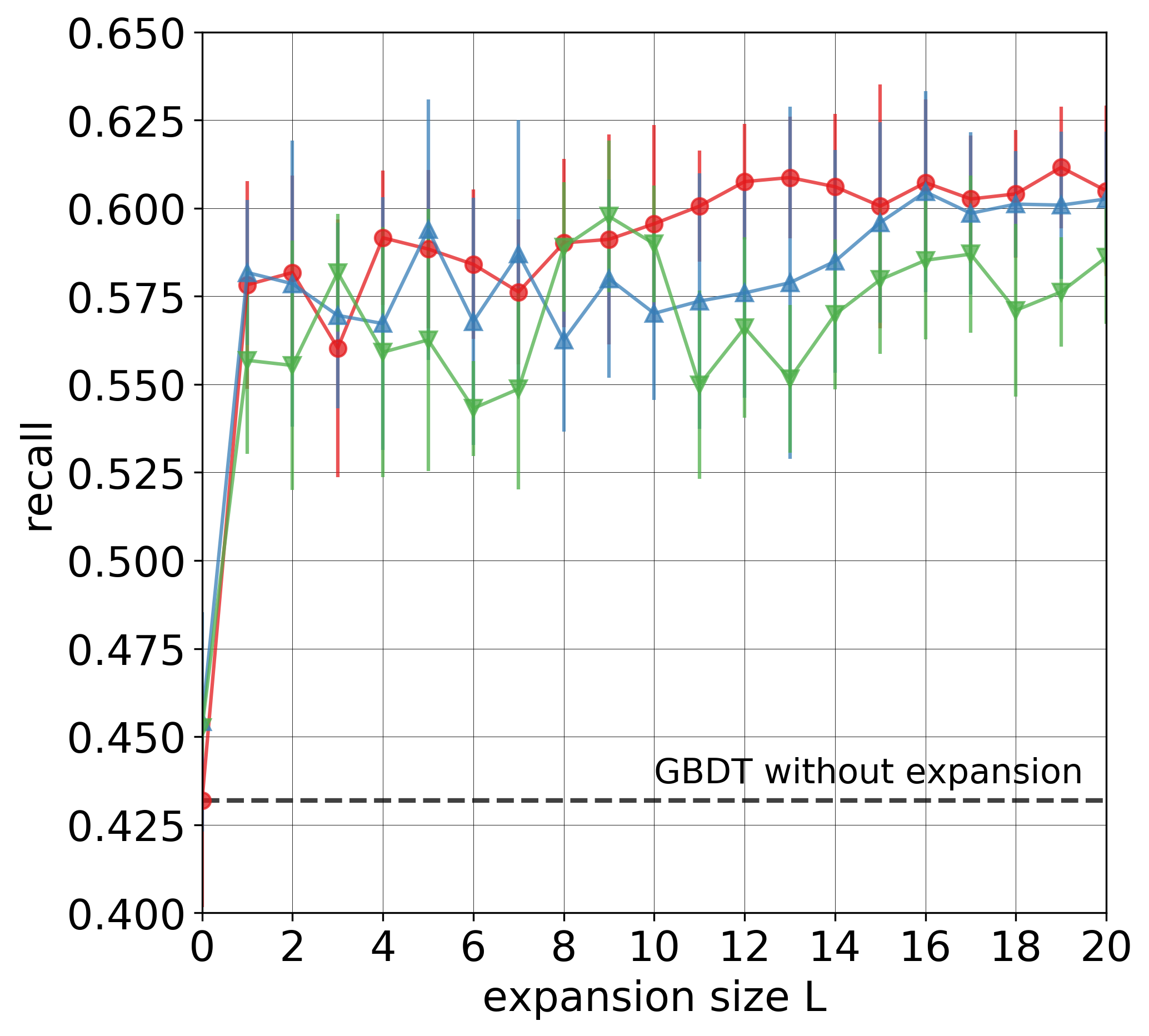}
        \caption{\textit{Recall}.}
        \label{fig:sel-recall}
    \end{subfigure}\hspace{0.04\textwidth}\begin{subfigure}[t]{0.27\textwidth}
        \centering
        \includegraphics[width=1\textwidth, angle=0]{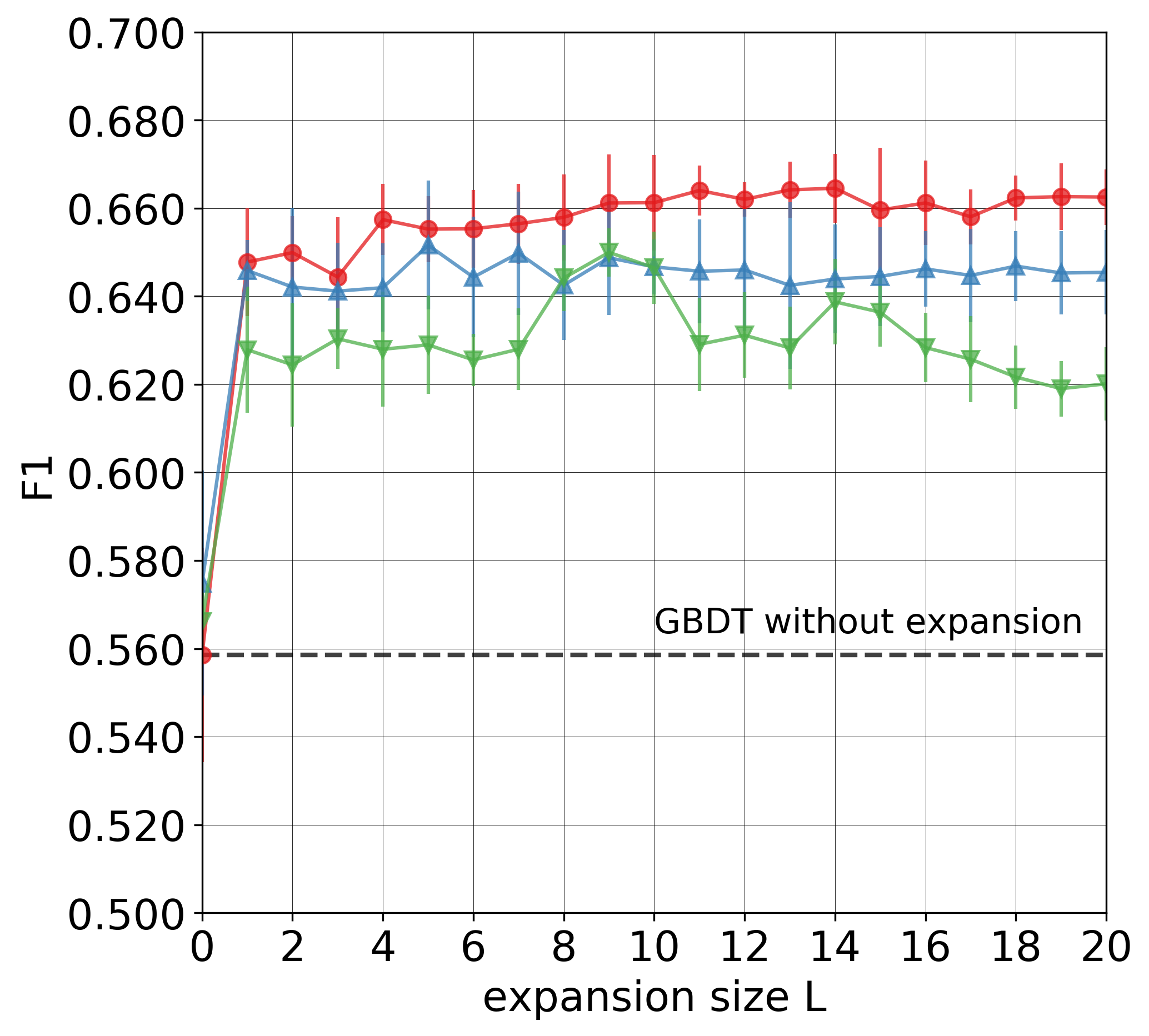}
        \caption{\textit{F1}.}
        \label{fig:sel-f1}
    \end{subfigure}\\\par\bigskip
    \begin{subfigure}[t]{0.27\textwidth}
        \centering
        \includegraphics[width=1\textwidth, angle=0]{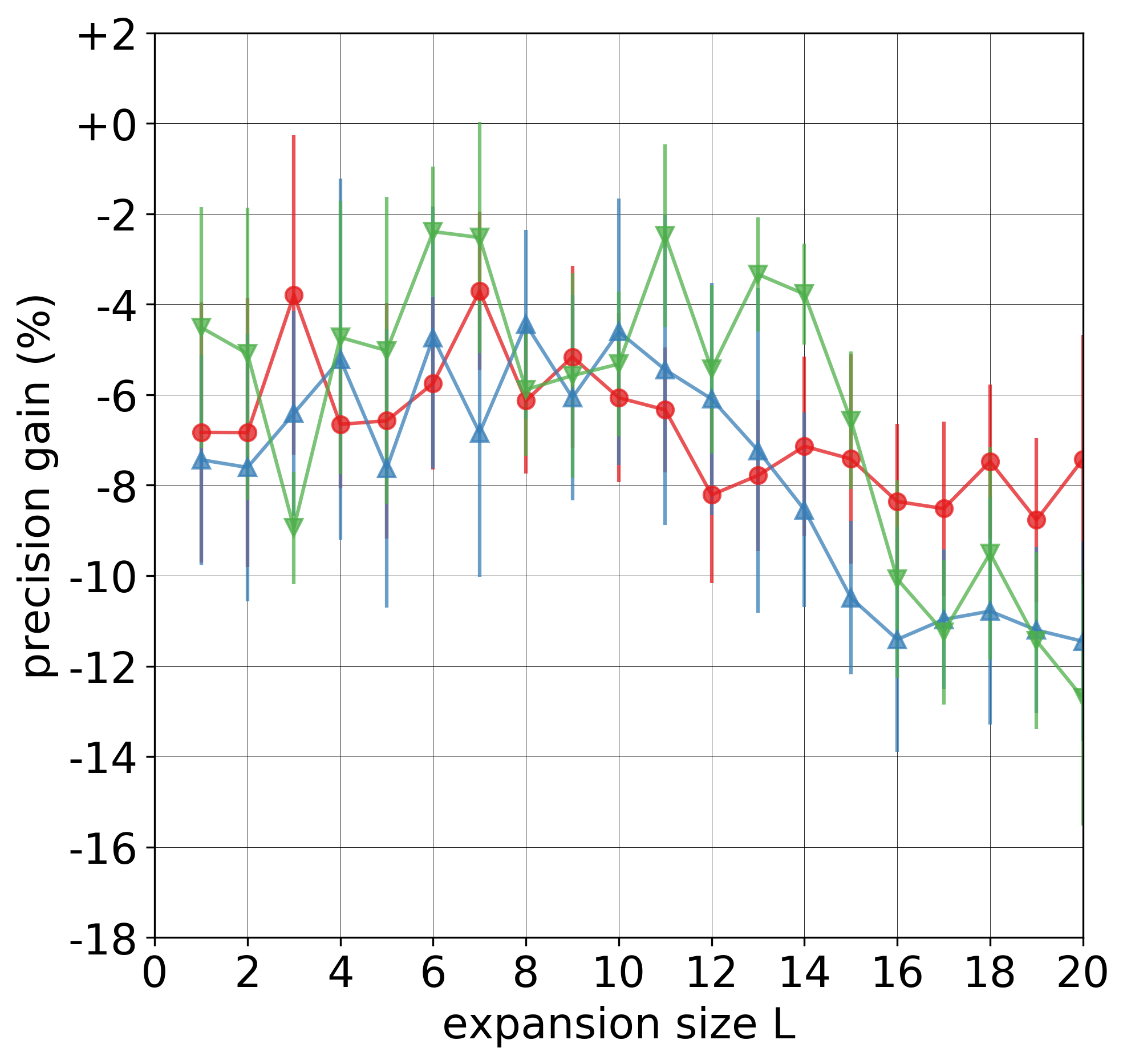}
        \caption{\textit{Precision} gain.}
        \label{fig:sel-precision-gain}
    \end{subfigure}\hspace{0.04\textwidth}\begin{subfigure}[t]{0.27\textwidth}
        \centering
        \includegraphics[width=1\textwidth, angle=0]{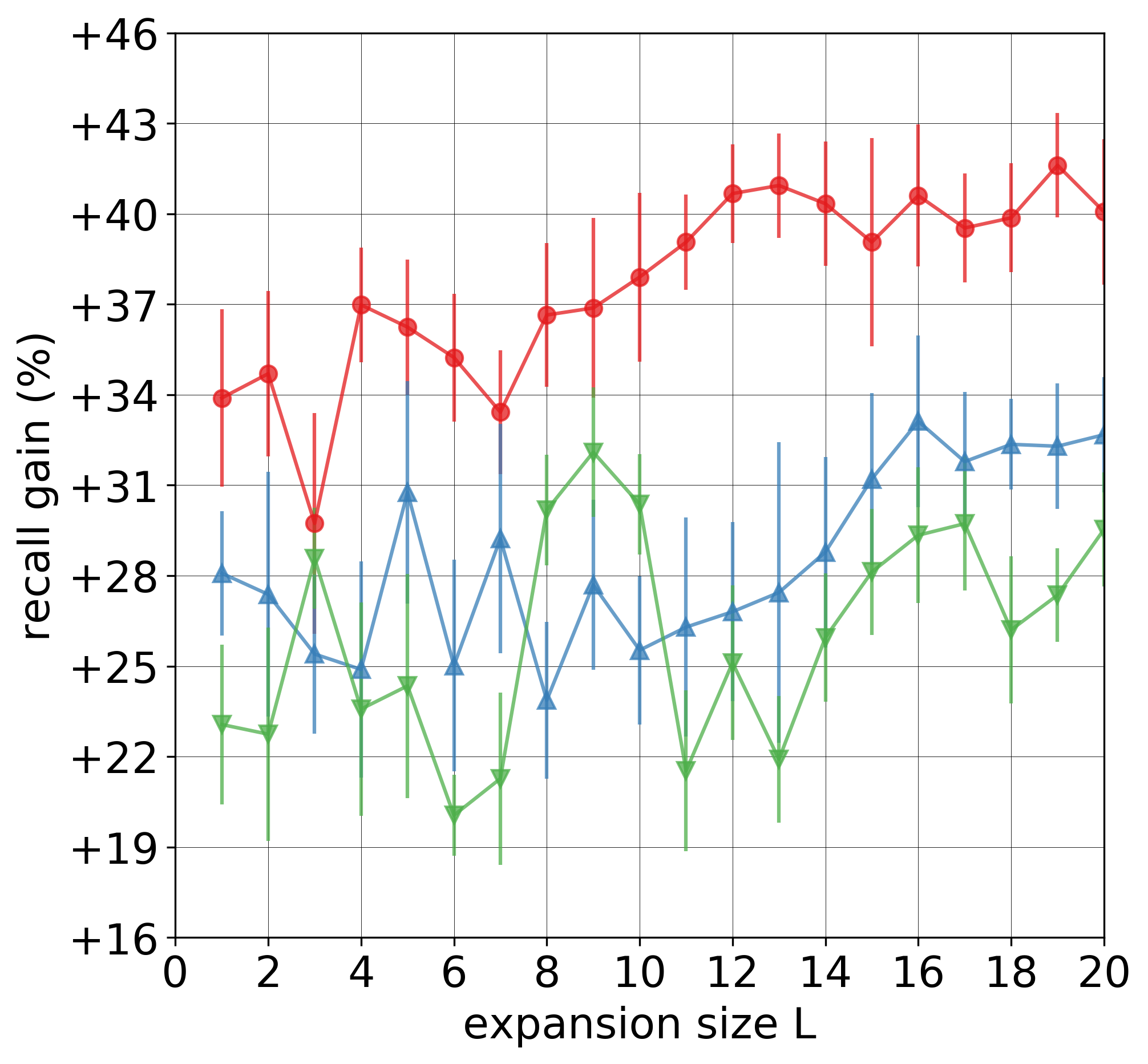}
        \caption{\textit{Recall} gain.}
        \label{fig:sel-recall-gain}
    \end{subfigure}\hspace{0.04\textwidth}\begin{subfigure}[t]{0.27\textwidth}
        \centering
        \includegraphics[width=1\textwidth, angle=0]{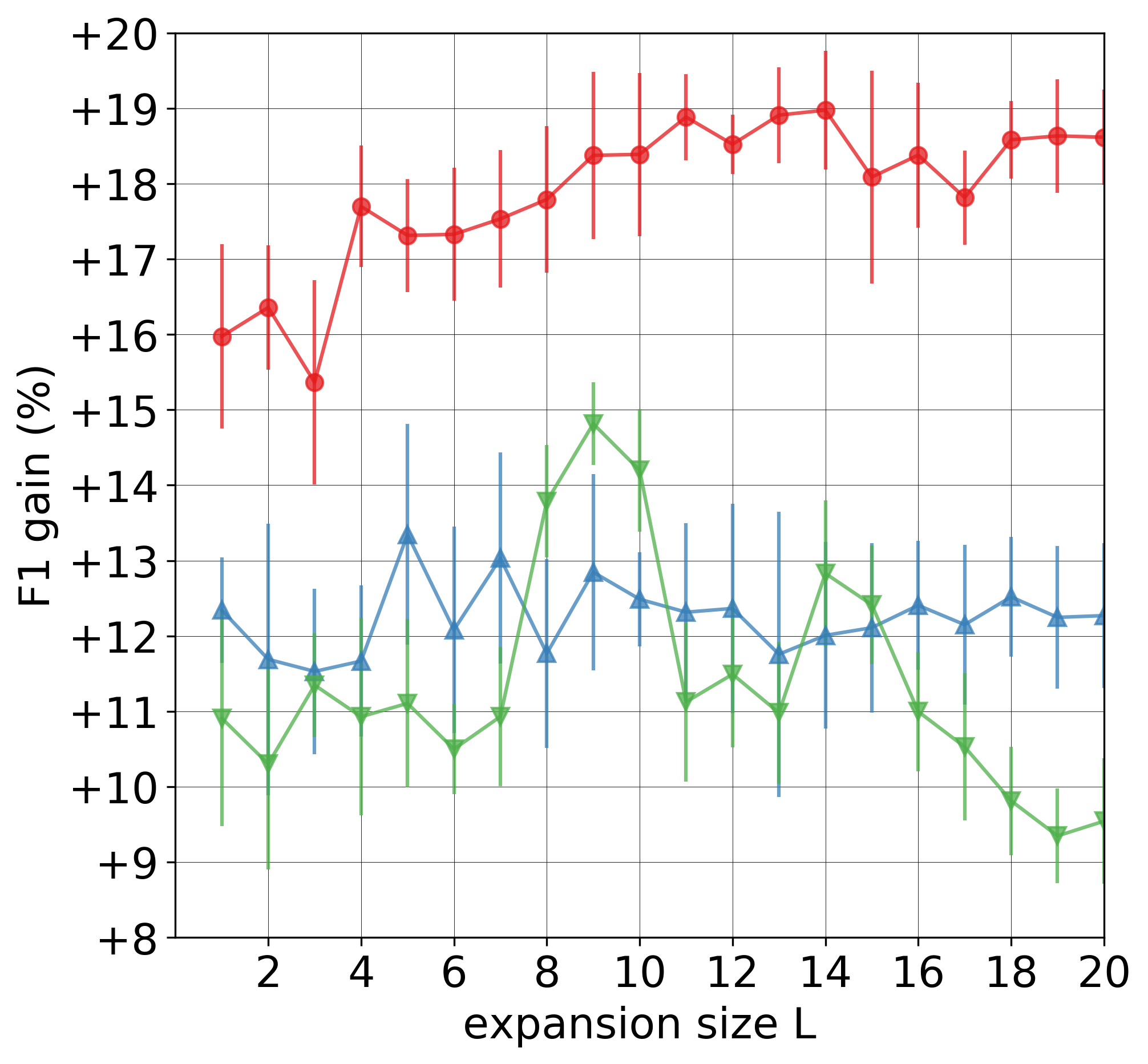}
        \caption{\textit{F1} gain.}
        \label{fig:sel-f1-gain}
    \end{subfigure}\caption{Performance comparison of different regression algorithms for the selection task, when varying the expansion size $L$.}
\label{fig:sel-results}
\end{figure*}

\paragraph{Results} In Figure~\ref{fig:sel-results}, we present a performance comparison of the different regression algorithms for the selection task, as a function of the expansion size $L$. In particular, Figures~\ref{fig:sel-precision},~\ref{fig:sel-recall} and~\ref{fig:sel-f1} report the trends of the evaluation metrics for $0 \leq L \leq 20$. Instead, Figures~\ref{fig:sel-precision-gain},~\ref{fig:sel-recall-gain} and~\ref{fig:sel-f1-gain} provide the percentage gain/loss attributable to the expansion step (thus, for $L>0$) with respect to the case without any expansion ($L=0$). Each point in the figures represents the mean scores obtained on the 10 runs, while error bars represent the corresponding standard deviation.

A first consideration regards the whole \texttt{GSP} approach. As expected, the expansion and selection steps provide complementary contributions. In fact, as the expansion size $L$ grows, so does the overall geoparsing \textit{recall}, as visible in figures~\ref{fig:sel-recall} and~\ref{fig:sel-recall-gain}. Depending on the algorithm and the expansion size, the improvement in recall ranges from $+21\%$ to $+42\%$. This result confirms our starting motivation of using expansion for enriching the set of candidate entities from which to extract pertinent geographic information. It also confirms our previous findings related to the \textit{maximum theoretical recall} of the different expansion strategies. However, as shown in figures~\ref{fig:sel-precision} and~\ref{fig:sel-precision-gain}, expanding also hinders \textit{precision}, given the higher likelihood of false positive predictions. Hence the need for an accurate selection step to mitigate losses in precision. In our experiments, such losses range from $-2\%$ to $-12\%$. As a consequence of these results, the overall balance achieved by \texttt{GSP} is very positive, meaning that the striking recall gain largely outweighs the relatively small precision loss. This is demonstrated by the \textit{F1} trends, presented in figures~\ref{fig:sel-f1} and~\ref{fig:sel-f1-gain}. Although the previous considerations hold for all regression algorithms, 
GBDT achieves consistently better results with respect to RF and DART. In particular, GBDT achieves the best global result with $F1=0.665$ (resulting in a $+19\%$ \textit{F1} gain) at the expansion size $L=14$. This result corresponds to \textit{precision} $=0.737$ ($-7\%$) and \textit{recall} $=0.606$ ($+40\%$), confirming our previous point.

Finally, Figure~\ref{fig:sel-f1} highlights the existence of a performance plateaux for $L\geq4$, revealing a sort of ``saturation'' in the learning process at larger expansion sizes. This effect may be due to the shortage of training examples for which good candidates are retrieved only at large expansion sizes. In turn, this shortage of training examples prevents our models from effectively learning how to recognize them. Larger annotated datasets may allow to delay this plateaux, resulting in additional performance improvements for large expansion sizes.

Building on our results so far, from now on we consider \texttt{GSP} with its best configuration resulting from the adoption of topological expansion with spelling-based sorting and the GBDT algorithm for the selection step. Expansion size is set to $L=14$.
 \section{Geoparsing results}
\label{sec:results}
In this section we advance our thorough analysis of \texttt{GSP}'s geoparsing results. Firstly, we present the details of our dataset. Then, we compare the best configuration of \texttt{GSP} with several baselines, with 2 state-of-the-art geoparsing algorithms, and with our previous technique. Finally, we provide additional results on \texttt{GSP}'s predictions, both in terms of identifying the most informative features for the selection task and of assessing the spatial granularity of our predictions.

\subsection{Dataset}
\label{sec:res-dataset}
In Section~\ref{sec:intro}, we underlined the importance of the geoparsing task to properly integrate OSN data in decision support systems. Although geoparsing techniques can process any kind of texts (e.g., news articles, emails), we remarked the challenges posed by OSN user-generated content, which is characterized by short texts, poor context and use of jargon and colloquial expressions. For these reasons, we train and evaluate our technique on a dataset composed of OSN user-generated posts. 

\begin{figure*}[t]
    \centering
    \begin{subfigure}[t]{0.27\textwidth}
        \centering
        \includegraphics[width=1\textwidth]{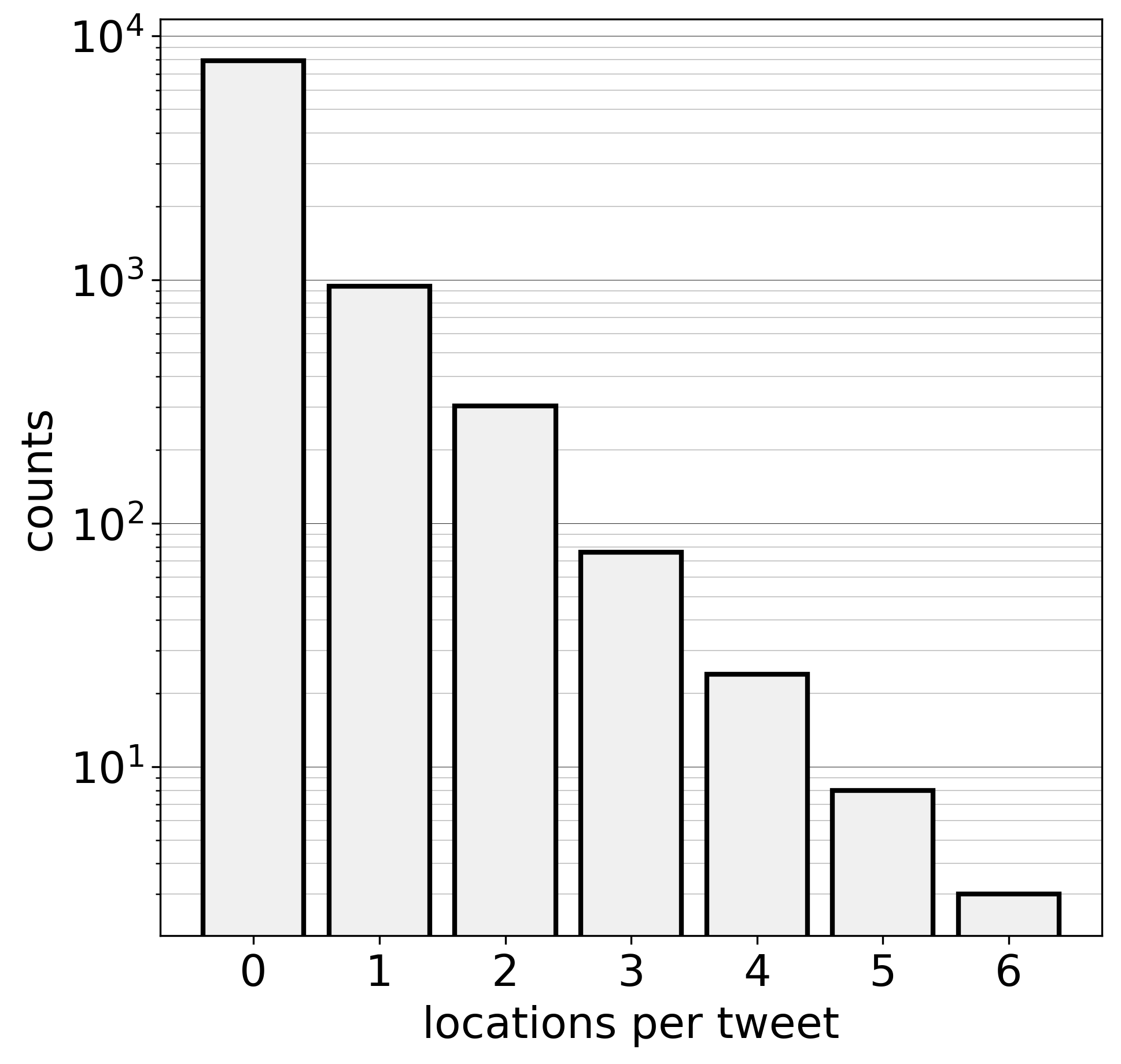}
        \caption{Distribution of mentioned locations per tweet.}
\label{fig:tweet_loc_dist}
    \end{subfigure}\hspace{0.05\textwidth}\begin{subfigure}[t]{0.5\textwidth}
        \centering
        \includegraphics[width=1\textwidth]{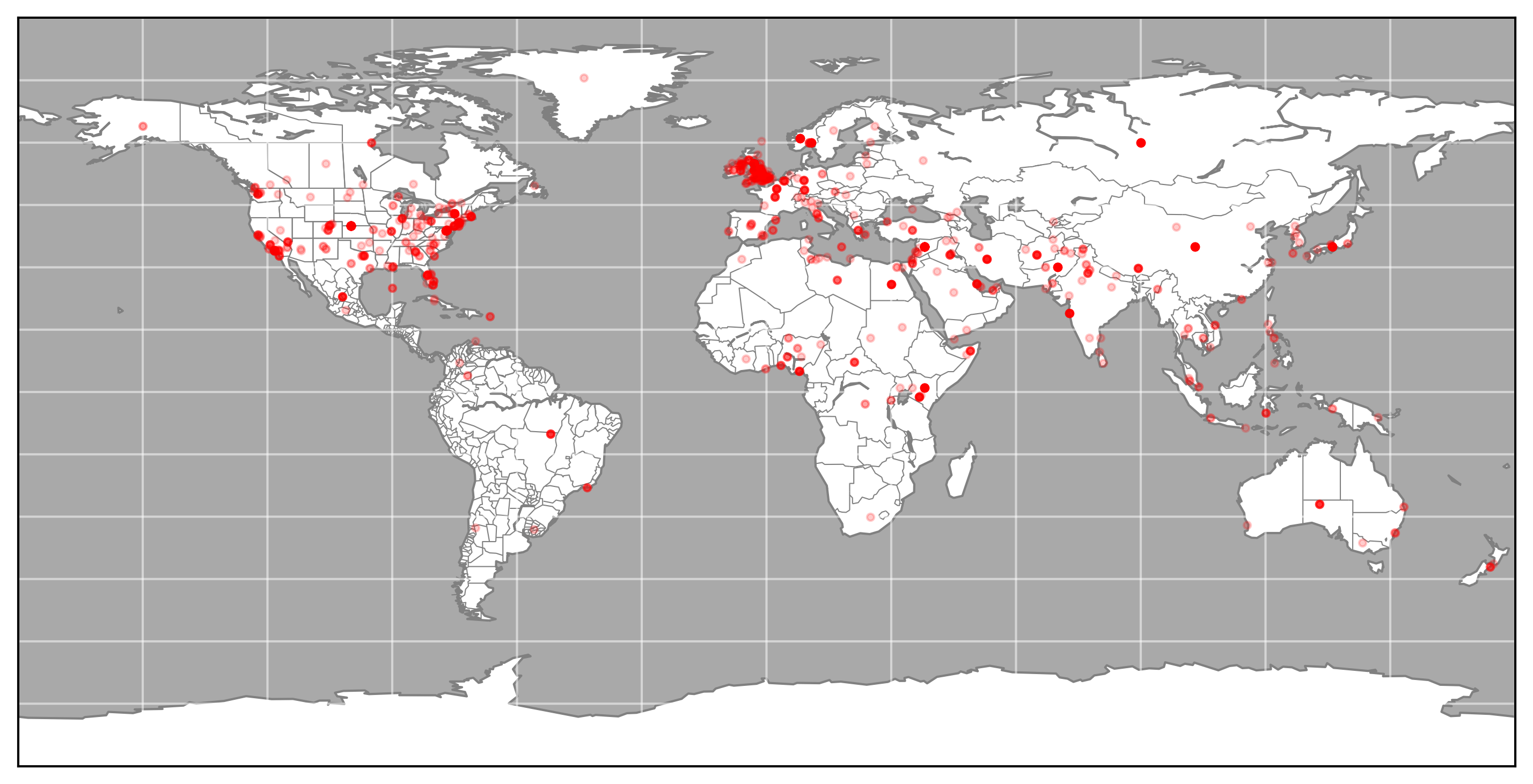}
        \caption{Geographic distribution of ground truth locations.}
\label{fig:loc_map}
    \end{subfigure}\caption{\textit{NEEL16} dataset profiling.}
    \label{fig:dataset_dist}
\end{figure*}

In particular, we use the official dataset of the \textit{2016 Named entity recognition and linking} challenge (\textit{NEEL16})\footnote{\scriptsize\url{https://aclweb.org/portal/content/named-entity-recognition-and-linking-challenge}}. This well-known, reference dataset includes 9,289 English tweets, extracted from a corpus of over 18M documents, covering several noteworthy events from 2011-13, and a set of hashtags from 2014-15. Notably, \textit{NEEL16} challenge organizers provide annotations about mentions of places/locations, enabling the usage of the dataset for geoparsing purposes. In particular, each ground truth location is complemented with the corresponding geographic coordinates as well as with the link to the corresponding resource on the English \textit{DBpedia}. The total number of locations is 5,348. Their distribution across tweets is very skewed, as shown in Figure~\ref{fig:tweet_loc_dist}. Namely, 85.4\% of the tweets do not mention any location, 10.1\% of the tweets mentions one location, and only 4.5\% of the tweets mentions multiple locations. Notably, locations are scattered all over the world, as shown in Figure~\ref{fig:loc_map}. Working at the world scale poses a severe challenge to geoparsing methods, but at the same time it is a requirement for many real-world applications. As a result, the \textit{NEEL16} dataset provides a suitable playground for training and evaluating geoparsing techniques.

Starting from the complete \textit{NEEL16} dataset, we obtain training (64\% of tweets), validation (16\%) and test (20\%) sets by performing a stratified sampling over the number of locations per tweet. We also include tweets without any mention of locations, because we want to ensure that all evaluated techniques do not return false positive predictions for them. As a result of this sampling strategy, the total number of locations is balanced across the obtained dataset splits. Moreover, performing the split at the message level -- and not at the single location instance level -- ensures that instances in the test set cannot affect the model training and validation. In this way, we evaluate our models with a fully blind test, correctly assessing possible overfitting. Our evaluation for this work is more severe than the one used in our previous work~\cite{avvenuti2018gsp}, thus explaining the differences in the reported performance.

\subsection{Performance comparisons}
\label{sec:res-comparisons}
\paragraph{Benchmarks} Performance comparisons are aimed at evaluating the performance of our proposed \texttt{GSP} technique, with reference to those of baselines and other advanced geoparsing systems. The first baseline that we implemented (labeled na\"{i}ve geoparser), leverages the \textit{geopy} Python package\footnote{\scriptsize\url{https://geopy.readthedocs.io/}} as an interface to the online \textit{ArcGIS} geocoding service\footnote{\scriptsize\url{https://developers.arcgis.com/features/geocoding/}}. This service is designed to geocode well-formatted addresses, but it also includes a simple free-text processing feature. As such, it represents a rather simplistic approach 
for geoparsing short and noisy documents. The second baseline (labeled NER+geocoder) is a basic implementation of the NER+gazetteer lookup approach. We implement the NER step using the well-known \textit{polyglot} natural language processing framework\footnote{\scriptsize\url{https://polyglot.readthedocs.io/}}. Then, we perform the gazetteer lookup step by means of the \textit{Google Maps} geocoder.
Adding to the 2 simple baselines, we also compare our results against 2 state-of-the-art geoparsers. As anticipated in Section~\ref{sec:related}, we include as benchmarks the techniques proposed by Middleton \textit{et al.}~\cite{middleton2014b} and by Halterman~\cite{halterman2017mordecai}. Both techniques leverage the common approach to geoparsing based on Named-Entity Recognition and geographic gazetteer lookup. Middleton \textit{et al.}'s technique returns the location tokens extracted from the input text, delegating the user to query the \textit{OpenStreetMap} gazetteer by means of a dedicated API service. Instead, Halterman's technique directly performs also the disambiguation step, returning the location coordinates. In fact, this model is more sophisticated than that by Middleton \textit{et al.}, leveraging a deep learning neural network classifier to select the proper instances in the \textit{GeoNames} gazetteer. In both cases, once the location named entities are linked to the proper gazetteer entries, the system returns the related coordinates.
Finally, we also include our earlier geoparsing technique~\cite{avvenuti2018gsp} in the comparison. Here, our goal is that of evaluating the effectiveness of the novel expansion and selection steps, with respect to the simpler approach developed in~\cite{avvenuti2018gsp}.

\paragraph{Evaluation metrics and experimental setup} We evaluate geoparsing techniques according to the \textit{precision}, \textit{recall} and \textit{F1} metrics, described in Section~\mbox{\ref{sec:problem}}. Moreover, we introduce also the \textit{elapsed time}, defined as the average time that a geoparsing technique needs to process a single tweet belonging to the test set. All the compared techniques are implemented with the Python language and deployed on a machine equipped with an 8-core \emph{CPU} featuring 50Gb of \emph{RAM}, and a Nvidia Tesla K80 GPU.

\begin{table}[t]
	\scriptsize
	\centering
	\begin{tabular}{lcrrcrrcrrcrr}
		\toprule
		&& \multicolumn{11}{c}{\textbf{evaluation metrics}} \\
		\cmidrule{3-13}
		\textbf{technique} && \multicolumn{2}{c}{\textit{precision ($\pm$\%)}} && \multicolumn{2}{c}{\textit{recall ($\pm$\%)}} && \multicolumn{2}{c}{\textit{F1 ($\pm$\%)}} && \multicolumn{2}{c}{\mbox{\textit{elapsed time ($\pm$\%)}}} \\
		\midrule
		\multicolumn{10}{l}{\textit{Benchmarks}} \\ [0.5em]
		na\"{i}ve geoparser                             && 0.033 & ($-$2133.33) && 0.157 & ($-$285.99) && 0.054 & ($-$1131.48) && 0.811 s & \mbox{($+$ 60.30)} \\
		NER+geocoder                                    && 0.300 & ($-$145.67) && 0.267 & ($-$126.97) && 0.282 & ($-$135.82) && 0.113 s & \mbox{($-$ 184.96)} \\
		Middleton \textit{et al.}~\cite{middleton2014b} && 0.123 & ($-$499.19) && 0.333 & ($-$81.98) && 0.180 & ($-$269.44) && \textbf{0.011 s} & \mbox{($-$ 2827.27)} \\
		Halterman~\cite{halterman2017mordecai}		    && 0.320 & ($-$130.31) && 0.299 & ($-$102.68) && 0.309 & ($-$115.21) && 0.049 s & \mbox{($-$ 557.14)} \\
		Avvenuti \textit{et al.}~\cite{avvenuti2018gsp} && \textbf{0.818} & ($+$9.90) && 0.417 & ($-$45.32) && 0.553 & ($-$20.25) && 0.321 s & \mbox{($-$ 0.31)} \\
		\midrule
		\multicolumn{10}{l}{\textit{Our contribution}} \\ [0.5em]
		\texttt{GSP}                                    && 0.737 &&& \textbf{0.606} &&& \textbf{0.665} &&& 0.322 s & \\
		\bottomrule
	\end{tabular}
	\caption{Geoparsing performance comparison between \texttt{GSP}, 2 baselines, and 3 state-of-the-art geoparsing techniques. Best results for each metric are shown in \textbf{bold} font. Beside each metric, we report in parentheses the \% gain/loss of each technique with respect to \texttt{GSP}. All differences between \texttt{GSP} and the other benchmarks are statistically significant, except for the elapsed time with respect to Avvenuti \textit{et al.}}
\label{tab:results}
\end{table} 
\paragraph{Results} Table~\ref{tab:results} reports a thorough comparison of geoparsing results. As shown, \texttt{GSP} outperforms all competitors. Both baselines performed rather poorly, as expected for a challenging task such as geoparsing. However, surprisingly the NER+geocoder baseline managed to beat the system in~\cite{middleton2014b}. An analysis of the results reveals that this is mainly due to the need to perform geoparsing at world-level, which made it challenging for~\cite{middleton2014b} to correctly disambiguate detected toponyms. As we discussed in Section~\ref{sec:related}, many geoparsing systems based on gazetteer lookup need to be constrained to operate in a geographically limited area, for maintaining satisfactory performance. In our case, this could not be done,
since the \textit{NEEL16} dataset is geographically unconstrained, 
including locations from all over the world, as shown in Figure~\ref{fig:loc_map}.

When compared to previous geoparsing techniques, the $F1$ gain of \texttt{GSP} ranges from $+269.44\%$ with respect to Middleton \textit{et al.}~\cite{middleton2014b} to $+20.25\%$ with respect to our previous technique~\cite{avvenuti2018gsp}. The improvement from our previous attempt at the geoparsing task and \texttt{GSP} is determined by our higher recall. Indeed, a relatively low recall was the limiting factor in~\cite{avvenuti2018gsp}. Having improved on the recall, our proposed \texttt{GSP} technique managed to obtain overall better results. In turn, this motivates our design choices related to the expansion step. Anyway, the large recall gain is partly counterbalanced by a slightly reduced precision ($-9.90\%$), demonstrating the difficulty at correctly selecting the best entity among those retrieved during the expansion step.

The price for more accurate predictions is slightly paid in terms of time efficiency. The last column of Table~\mbox{\ref{tab:results}} shows that \texttt{GSP} (both in its present and earlier version) needs around 0.3 seconds to geoparse a single tweet, compared to 0.05 of Halterman and 0.01 of Middleton \textit{et al}. Despite being computationally more demanding, \texttt{GSP} is nonetheless suitable for real-time applications, also considering the possibility to deploy multiple parallel instances in those cases requiring high throughput.

\subsection{Feature importance analysis}
\begin{figure*}[t]
    \centering
    \begin{subfigure}[t]{0.36\textwidth}
        \centering
        \includegraphics[width=1\textwidth]{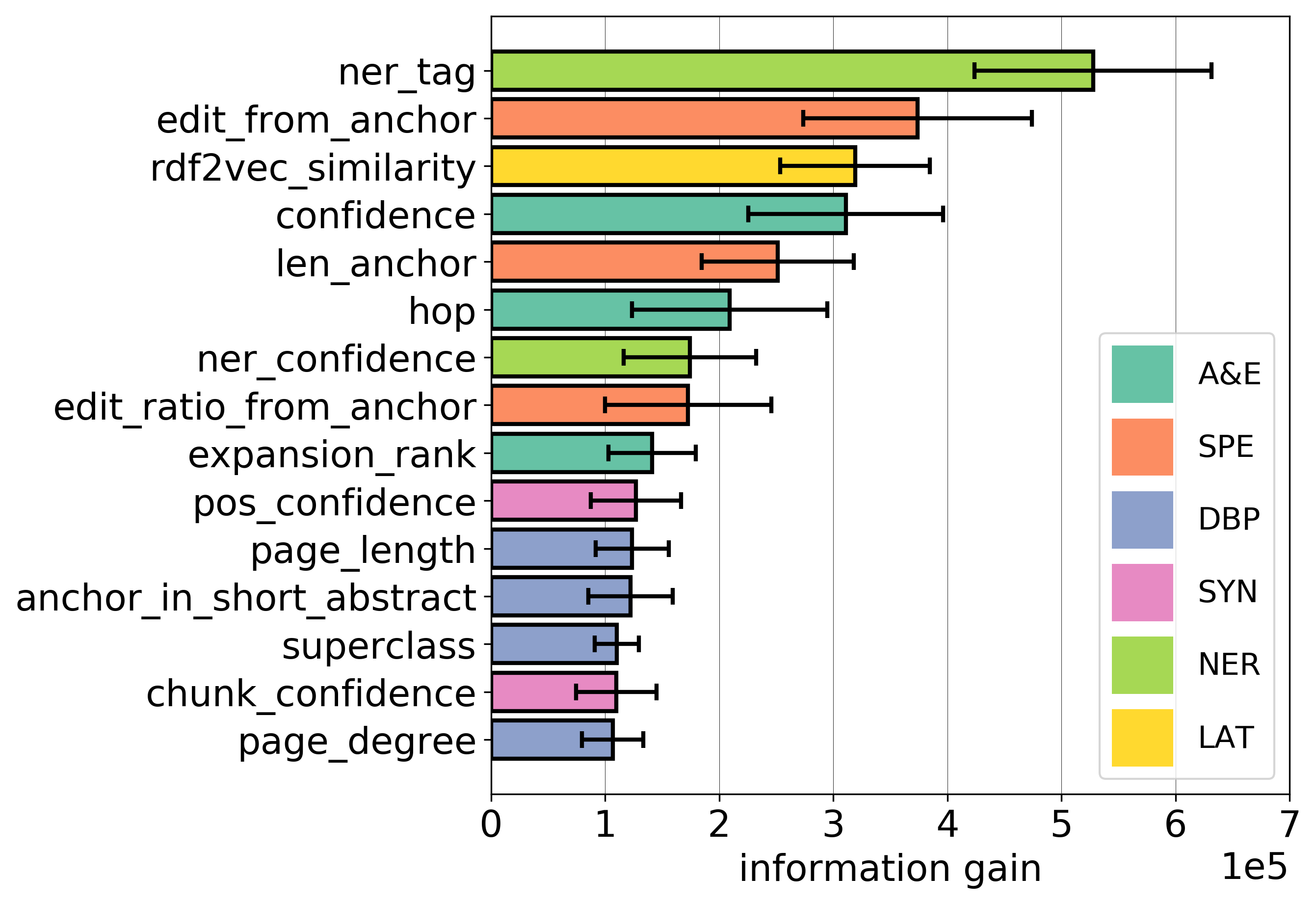}
        \caption{Individual feature importance.}
\label{fig:featimp}
    \end{subfigure}\hspace{0.05\textwidth}\begin{subfigure}[t]{0.26\textwidth}
        \centering
        \includegraphics[width=1\textwidth]{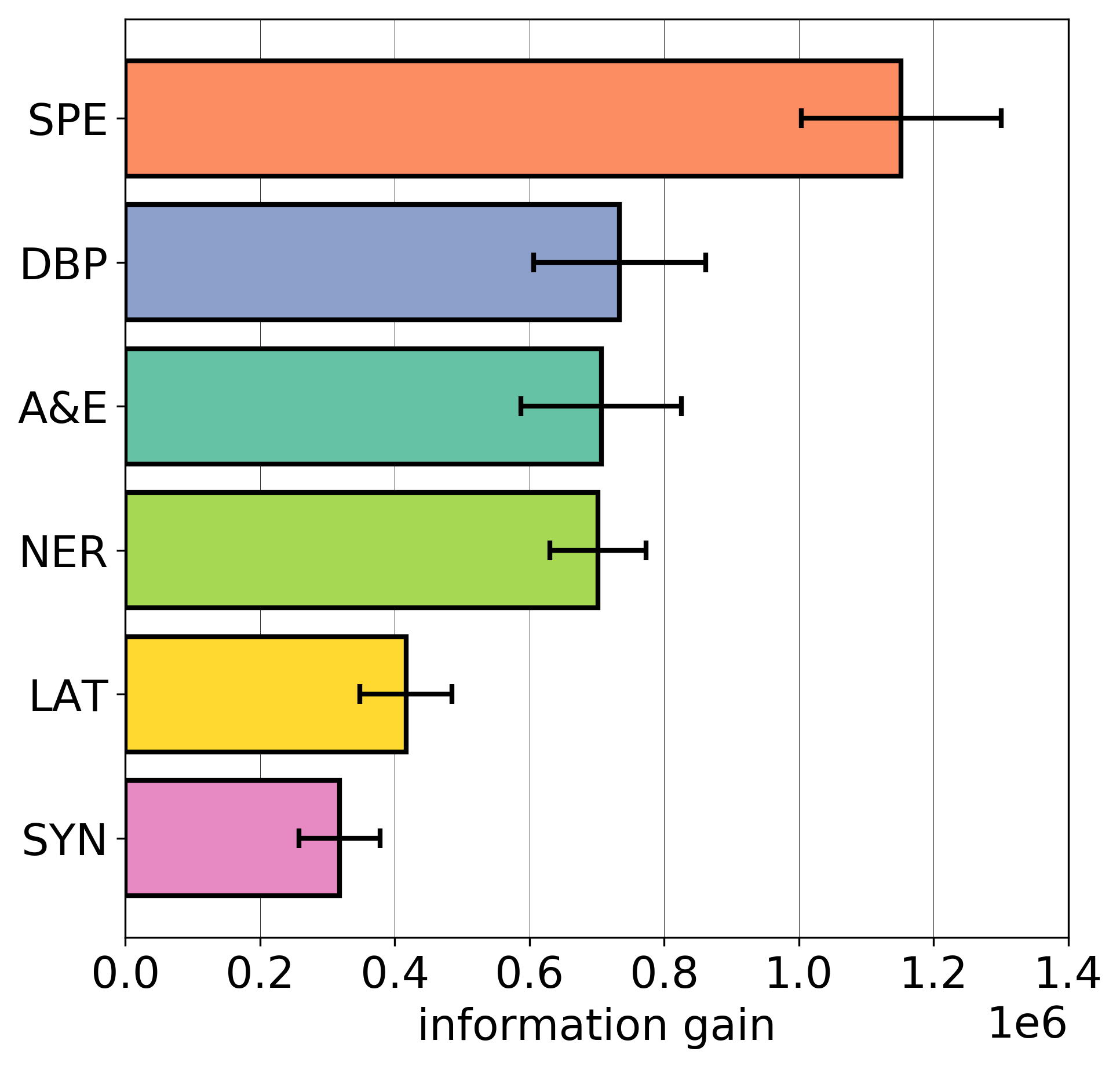}
        \caption{Feature importance by group.}
\label{fig:featimp_cat}
    \end{subfigure}\hspace{0.05\textwidth}\begin{subfigure}[t]{0.26\textwidth}
        \centering
        \includegraphics[width=1\textwidth]{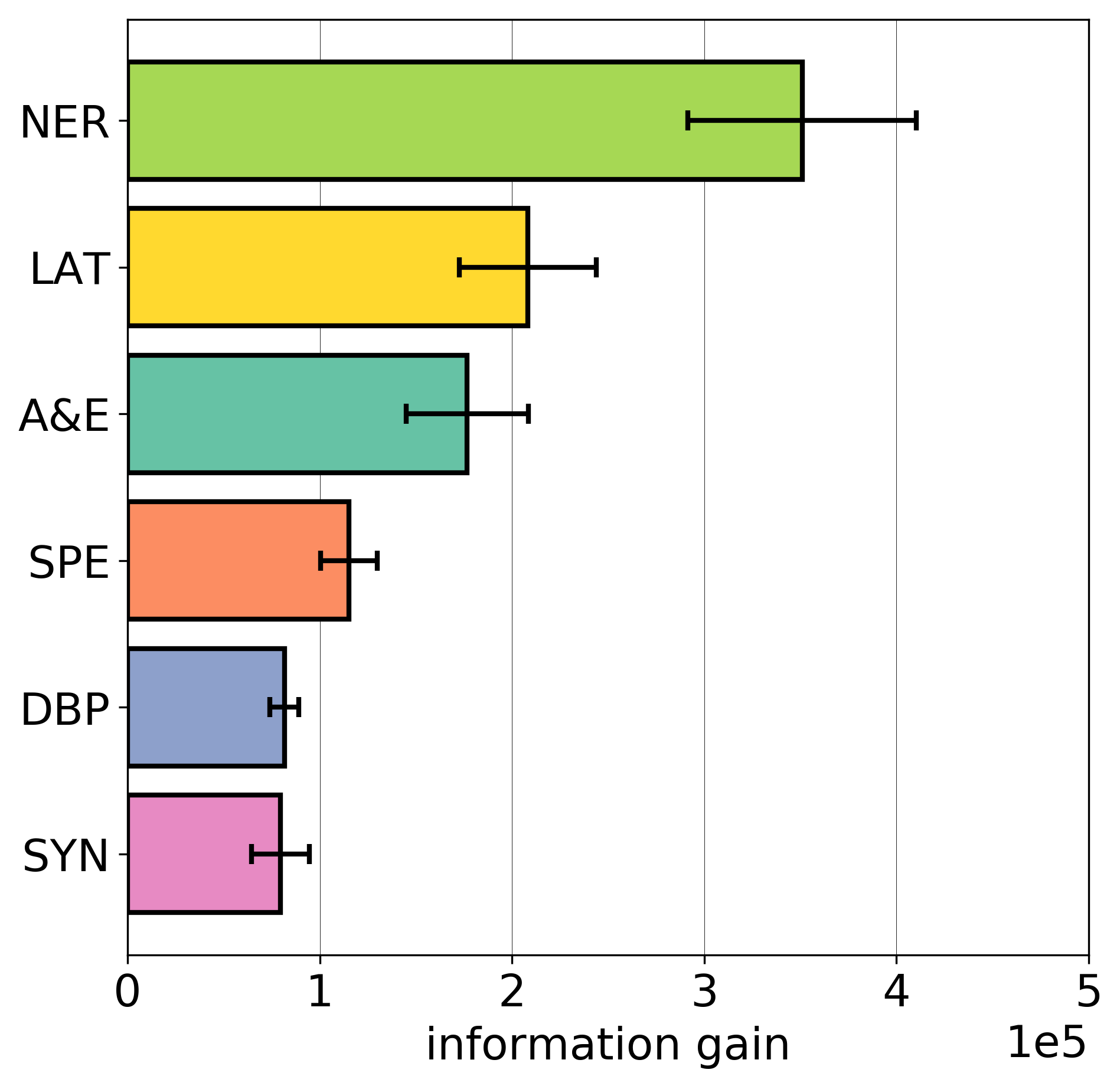}
        \caption{Feature importance by group, normalized by its cardinality.}
\label{fig:featimp_catnorm}
    \end{subfigure}\caption{Feature importance analysis for the best configuration of \texttt{GSP},
measured as the information gain provided by each feature in the regression model. We refer to Table~\mbox{\ref{tab:sel-features}} for details about the features and the related groups and acronyms.}
    \label{fig:feature_importance}
\end{figure*}

In Figure~\ref{fig:feature_importance} we report the results of the feature importance analysis for the best configuration of \texttt{GSP} (i.e., the one 
using GBDT for the selection step). Adopting a standard approach, we compute the feature importance as the sum of the \textit{information gain} provided by a feature each time it triggers a split in one of the trees composing the model. In particular, Figure~\ref{fig:featimp} shows the importance of the top-15 individual features. Interestingly, all feature 
groups make it to the top-15, meaning that each 
group conveys useful information for the model. The named-entity recognition (NER) tag is the most important feature in our model. This further confirms the validity of previous approaches to geoparsing, based on NER tagging and gazetteer lookup. As expected, also features measuring the difference between the anchor and the entity name play an important role (\textsf{edit\_from\_anchor} and \textsf{edit\_ratio\_from\_anchor}). The third most-informative feature is a latent semantic feature -- namely, the one computed out of \textit{rdf2vec} node embeddings. This feature measures the semantic difference between the starting node returned by the semantic annotator and the candidate obtained via expansion.

By leveraging information gain's \textit{additivity}, we also collectively evaluate feature importance by group. Figure~\ref{fig:featimp_cat} reports the results of this analysis, showing that \textit{spelling} (\texttt{SPE}) and \textit{DBpedia} (\texttt{DBP}) features account for the majority of contributions. This result confirms the importance of modeling the intrinsic properties and the similarities of the anchors and the entities. In order to avoid possible biases due to the numerosity of certain feature 
groups with respect to others, we also normalize the feature importance of each 
group by its cardinality. After normalization we observe a different ranking, dominated by the more sophisticated, information-rich \texttt{NER} and by latent semantic features (\texttt{LAT}). Instead, \textit{syntactic} (\texttt{SYN}) features seems to provide the smallest contribution throughout all experiments. We underline that these observations are intended to shed light on the overall learning process. In fact, a rigorous assessment of the feature importance should account for possible correlations between features, that is beyond the scope of this analysis.

\subsection{Place granularity analysis}
A favorable feature of \texttt{GSP} is the possibility to leverage knowledge graphs and the Linked Data ontology to infer the granularity of the predicted locations. Following previous works~\cite{middleton2014b}, we define 4 granularity levels: (i) points of interest (POIs), roughly corresponding to buildings and other notable landmarks; (ii) cities; (iii) regions and counties; and (iv) countries. To infer the granularity level of a given prediction by \texttt{GSP}, we analyze the \textit{DBpedia} ontology of the geographic entity providing the coordinates. For example, if we geotagged a token with the entity \texttt{Bath}, our prediction would be at the city-level, since the \textit{DBpedia} resource for Bath has \texttt{rdf:Type}$=$\texttt{dbo:City}\footnote{\scriptsize\url{http://dbpedia.org/page/Bath,_Somerset}}. By following a similar approach, we are able to assess the granularity of each distinct ground truth instance in the \textit{NEEL16} dataset, since ground truth annotations are complemented with \textit{DBpedia} URLs. We can thus perform a granularity-aware evaluation of \texttt{GSP}, by considering as true positives only those predictions matching both the ground truth coordinates and the corresponding granularity level.

\begin{figure*}[t]
    \centering
    \begin{subfigure}[t]{0.27\textwidth}
        \centering
        \includegraphics[width=1\textwidth]{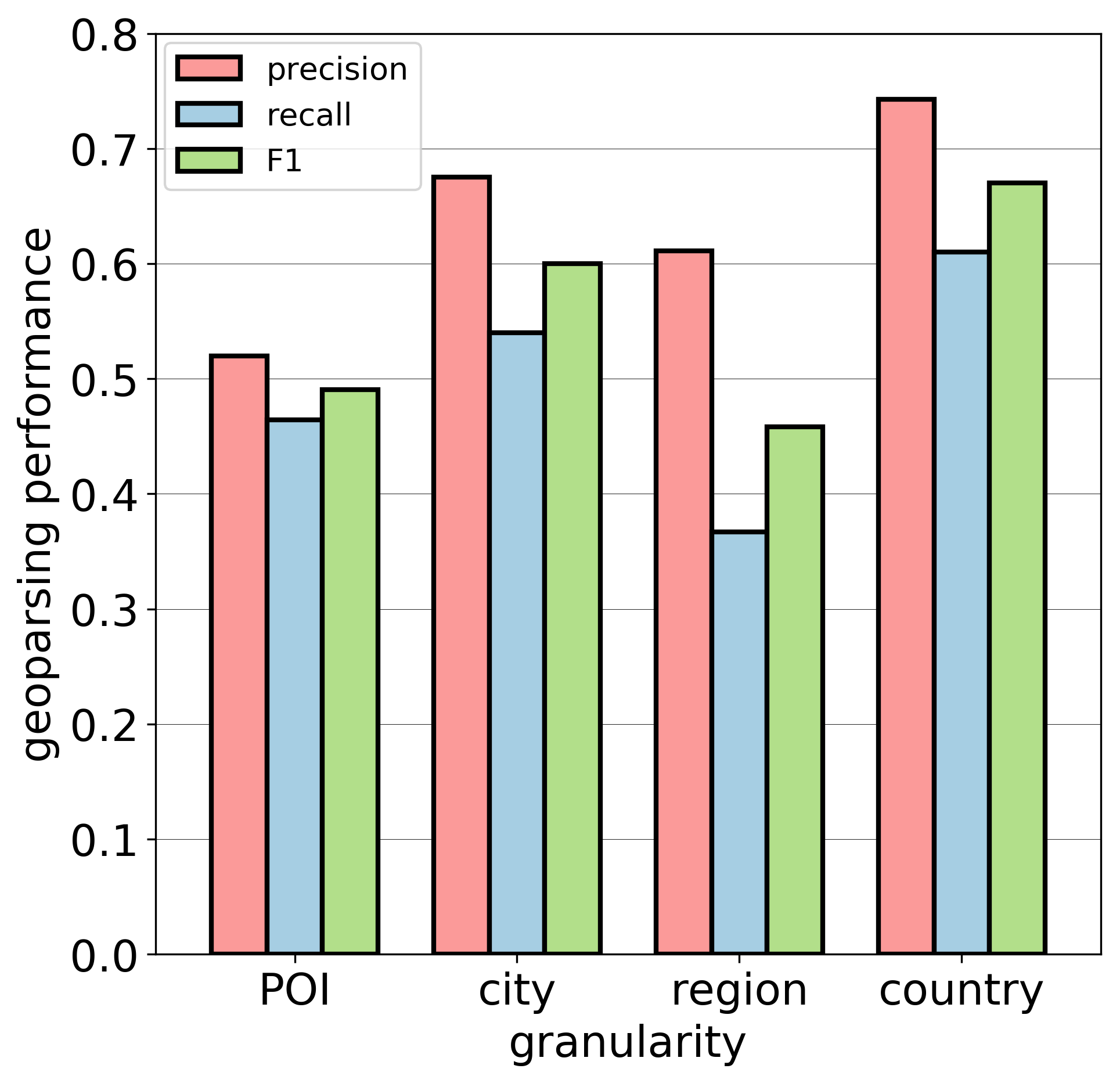}
        \caption{Geoparsing results of \texttt{GSP} at different granularity levels.}
        \label{fig:pred_gran}
    \end{subfigure}\hspace{0.15\textwidth}\begin{subfigure}[t]{0.27\textwidth}
        \centering
        \includegraphics[width=1\textwidth]{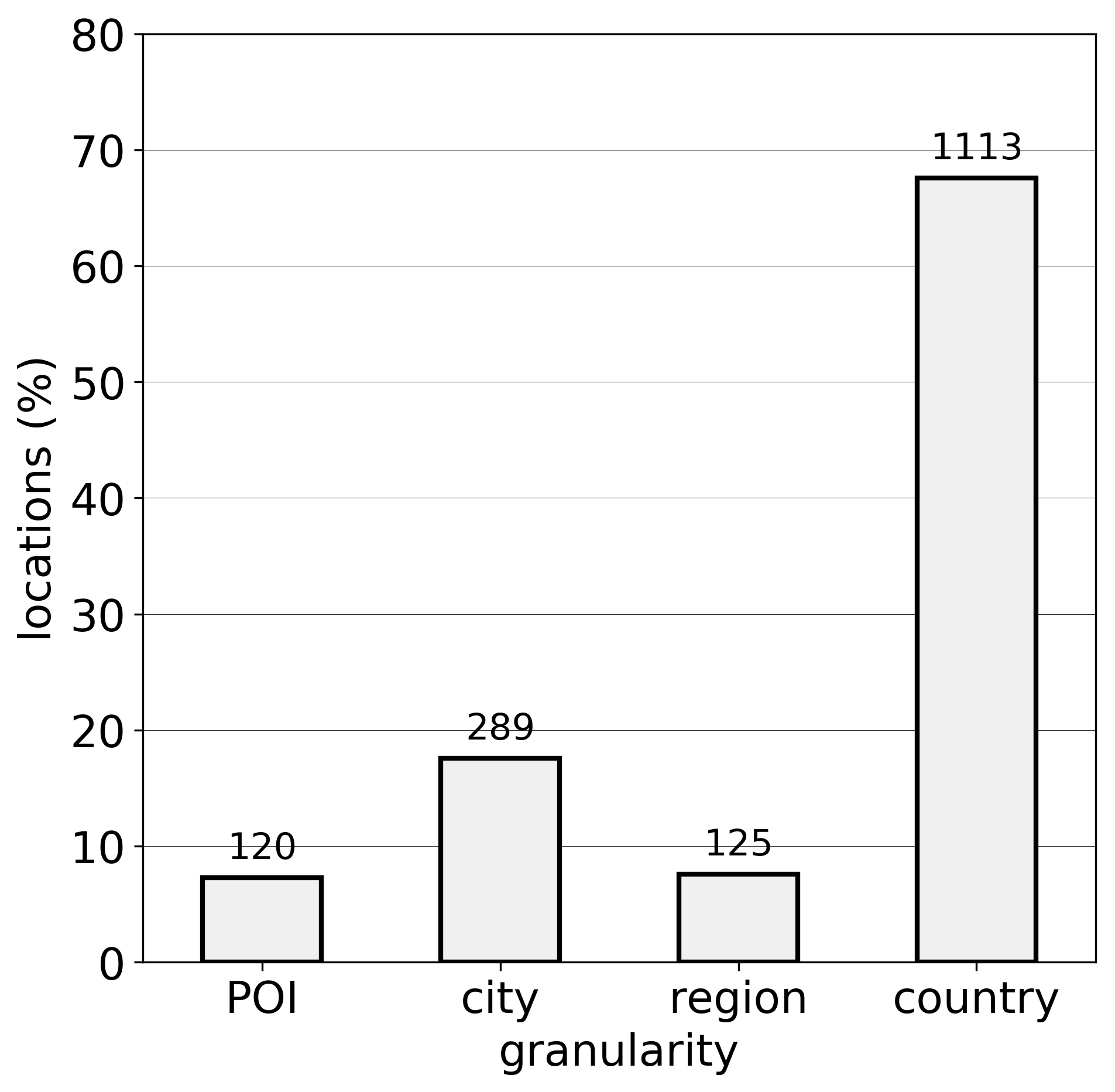}
        \caption{Distribution of distinct ground truth locations per granularity level.}
        \label{fig:gt_gran}
    \end{subfigure}\caption{Breakdown of \texttt{GSP} geoparsing results at different spatial granularity levels.}
\label{fig:selection_results}
\end{figure*}

We show the results of this more severe evaluation in Figure~\ref{fig:pred_gran}. The overall results are still satisfactory, with good performance at the country ($F1=0.670$) and city ($F1=0.600$) granularity levels. Conversely, the performance slightly drops for regions ($F1=0.458$) and POIs ($F1=0.491$). To explain this result, in Figure~\ref{fig:gt_gran} we provide the distribution of the ground truth granularity levels in the dataset. As shown, regions and POIs are significantly underrepresented. This possibly explains why our model struggled to learn instances at these granularity levels. In summary, \texttt{GSP} shows suitable performance in location granularity prediction, although there is room to improve it by enriching and balancing the training dataset.
 \section{Discussion}
\label{sec:discussion}
The geoparsing results of the proposed \texttt{GSP} technique, and the comparison with baselines and other state-of-the-art techniques, demonstrated the effectiveness of our design choices, and the favorable compelling performance of \texttt{GSP}.
Building on these results, in this section we discuss some additional features of our technique, with a specific focus on its \textit{robustness}, \textit{generalizability}, \textit{extensibility} and \textit{applicability}.

\subsection{Robustness and generalizability}
\label{sec:disc-robust}
Although evaluated on tweets, our technique does not make any assumption on the input text, and it does not exploit any peculiar feature of Twitter nor of OSNs in general. Because of this, it is suitable to geoparse any textual document, including longer texts such as news articles, emails and blog posts. The choice of evaluating our technique on the OSN-derived \textit{NEEL16} dataset stems from the will to test \texttt{GSP} on challenging texts. In fact, OSN user-generated content -- and \textit{tweets} specifically -- are known for their shortness, lexical sparsity, and for the use of abbreviations, jargon and colloquial expression. As such, they represent a proving ground for any text mining technique. Given this scenario, our already promising results are likely to further improve, should \texttt{GSP} be applied to geoparse longer and well-written texts. In addition to the challenges related to the analysis of short OSN texts, the \textit{NEEL16} dataset also presents other pitfalls. Indeed, it considers multiple different events and topics, spread across a large geographic area (as shown in Figure~\ref{fig:loc_map}), and encompassing several years. 

In conclusion, the positive results obtained on this challenging and diverse evaluation dataset guarantee that \texttt{GSP} can generalize well also to other texts and topics, thus proving its robustness and generalizability. Results of the application of \texttt{GSP} in-the-wild are thus likely to remain very positive. As a final remark on robustness and generalizability, we report that both \texttt{GSP} and its ancestor~\mbox{\cite{avvenuti2018gsp}} correctly geoparse the challenging \textit{toy example} of Section~\mbox{\ref{sec:expansion}}, whereas only the NER+geocoder technique succeeds among the other benchmarks.

\subsection{Extensibility}
\label{sec:disc-extensib}
Contrarily to other geoparsing techniques, \texttt{GSP} does not suffer from the restriction to be applied to geographically-limited or predefined areas. In fact, it can easily predict locations worldwide, as demonstrated in our experiments. However, the model 
proposed in this work was developed for processing English texts. This limitation does not derive from our design choices, that are totally language-independent, but it only depends on the availability of the language-specific components used by \texttt{GSP}. In other words, our technique could be applied to documents in any language, without any modification, provided that certain resources for that language exist. In particular, the first language-dependent component used by \texttt{GSP} is the semantic annotator. At the time of writing, the one used in our work (i.e., TagMe) processes English, German and Italian texts. However, other well-known annotators, such as DBpedia Spotlight\footnote{\scriptsize\url{https://www.dbpedia-spotlight.org/}}, support as much as 12 languages with the possibility to extend it to additional ones. 
Moreover, our technique needs language-specific models for NER, chunking and part-of-speech tagging, as well as for BERT and \textit{rdf2vec} embeddings. These requirements are similar to those of many other text mining techniques, and they can be easily met. In fact, many natural language processing (NLP) tools are available for the most widespread languages. As an example, the NLP library \textit{polyglot} supports from 16 to 196 languages, depending on the task. Similarly, BERT features a multi-lingual model, presently including 104 languages. Finally, an open-source library\footnote{\scriptsize\url{https://github.com/IBCNServices/pyRDF2Vec}} allows training \textit{rdf2vec} models for all existing \textit{DBpedia}s. This discussion highlighted which tools are needed to extend \texttt{GSP} to other languages, and where to start for building a deployment in a specific language.

\subsection{Applicability}
Thanks to its previously discussed robustness, generalizability and extensibility, the \texttt{GSP} technique proves suitable for integration in geo-spatial decision support systems based on OSN data, allowing very general and flexible settings. When empowered with \texttt{GSP}, those systems benefit from a significantly increased amount of accurate and structured geographic data, provided at the most specific granularity available. The comparison with state-of-the-art benchmarks, provided in Section~\mbox{\ref{sec:res-comparisons}}, demonstrated remarkably higher performance of \texttt{GSP}. In particular, methods based on NER+gazetteer lookup proved unable of working at the world scale, mainly due to their rough disambiguation approaches. This implies that such simple approaches can not be profitably used for tasks such as monitoring epidemic spreading or international tourism flows, contrarily to \texttt{GSP}.
Furthermore, although the higher complexity of \texttt{GSP} increases the time elapsed to process a message with respect to the other benchmarks, it still allows for practical, real-time applications.
Notably, the suitability of a simpler version of our geo-semantic-parsing technique for decision support systems was already proved in~\mbox{\cite{avvenuti2016impromptu}}. When integrated in the referenced crisis mapping system, \texttt{GSP} contributed to increase the fraction of georeferenced tweets from a poor 5\% to a remarkable 39\%, thus significantly extending the coverage and improving the accuracy of crisis maps.

 \section{Conclusions and future works}
\label{sec:conclusions}
Motivated by current limitations of existing geoparsing techniques, we proposed Geo-Semantic-Parsing (\texttt{GSP}) -- a novel technique for enriching text documents with structured geographic information. \texttt{GSP} leverages semantic annotation to identify relevant portions of the input text and to link them to pertinent entities in knowledge graphs, such as \textit{DBpedia}. Then, it exploits the information-rich and interconnected nature of the knowledge graph to retrieve additional entities from which to extract geographic information. To reach this goal, we devised an \textit{expansion} step that allows \texttt{GSP} to efficiently traverse the knowledge graph. Finally, in a dedicated \textit{selection} step, \texttt{GSP} selects the best entity with which to geotag the input document by solving a regression task. Extensive experimental results demonstrated the viability of our solution. In particular, \texttt{GSP} outperformed all state-of-the-art competitors achieving $F1=0.66$ versus $F1\leq0.55$ of other techniques. Due to its robustness, generalizability and extensibility, \texttt{GSP} can be integrated in geo-spatial decision support systems based on OSN data, empowering them with accurate and structured geographic data, available in real time. Notably, this kind of approach was able to increase the fraction of georeferenced tweets from a poor 5\% to a remarkable 39\% in a real-world crisis mapping setting.

Future works on geoparsing could investigate more sophisticated methods to effectively combine different, mutually-orthogonal expansion strategies, and possibly even multiple semantic annotators. As an alternative approach, interested stakeholders could also develop semantic annotators that are specifically designed to return pertinent geographic entities, given the importance of geographic information for many downstream tasks. Finally, an alluring line of research could involve investigating end-to-end geoparsing techniques, developed on top of state-of-the-art contextual word embeddings. In fact, word embeddings vector spaces are known to reflect spatial relationships between words in their topology. As a consequence, it could be possible to map the $N$-dimensional word embeddings vector space directly on the geographic space by learning a suitable projection function. Semantic knowledge graphs provide the ideal playground to learn such models, since they represent large corpora of textual documents. Moreover, such documents could be considered as already annotated, thanks to the hyperlink structure and the geographic information contained in the semantic resources.

\bibliography{bibliography}

\end{document}